\documentclass[a4paper]{article}

\usepackage{amsthm}
\usepackage{bm}
\usepackage{amsmath}
\usepackage{amssymb}
\usepackage{mathtools}
\usepackage{subfig}
\usepackage{xcolor}
\colorlet{RED}{red}
\usepackage{here}
\usepackage{natbib}
\usepackage{url}
\usepackage{authblk}
\usepackage{listings}
\usepackage{booktabs}
\usepackage{wrapfig}

\usepackage{algorithm}
\usepackage{algorithmic}

\usepackage{graphicx}
\newcommand{\red}[1]{\textcolor{red}{#1}}

\def\*#1{\boldsymbol{#1}}
\theoremstyle{plain}

\newcommand{\prm}[0]{\prime}
\newcommand{\tw}[0]{\textwidth}
\newcommand{\ig}[2]{\includegraphics[clip,width=#1\tw]{#2}}

\newcommand{\mr}[1]{\mathrm{#1}}

\newcommand{\argmax}{\mathop{\text{argmax}}}

\newcommand{\eq}[1]{(\ref{#1})}

\newcommand{\lw}[1]{\smash{\lower2.ex\hbox{#1}}}

\newcommand{\rbr}[1]{\left(#1\right)}
\newcommand{\sbr}[1]{\left[#1\right]}
\newcommand{\cbr}[1]{\left\{#1\right\}}

\newcommand{\RR}{\mathbb{R}}

\newcommand{\EE}{\mathbb{E}}

\newcommand{\cA}{{\cal A}}

\newcommand{\cC}{{\cal C}}
\newcommand{\cD}{{\cal D}}

\newcommand{\cF}{{\cal F}}
\newcommand{\cG}{{\cal G}}

\newcommand{\cL}{{\cal L}}

\newcommand{\cN}{{\cal N}}

\newcommand{\cP}{{\cal P}}

\newcommand{\cX}{{\cal X}}

\usepackage{geometry}

\ifdefined\draft
\newcommand{\note}[1]{}
\newcommand{\remove}[1]{}
\else
\newcommand{\note}[1]{\red{(Memo: #1)}}
\newcommand{\remove}[1]{%
\begingroup
\color{blue}
#1
\endgroup
}
\fi

\def\arxiv{}
\def\ackn{}
\newif\iftwocol
\twocolfalse

\title{
A Unified Information-Theoretic Approach to Constrained Multi-Fidelity Multi-Objective Bayesian Optimization
}
\author[1]{Rikuto~Matsumoto}
\author[1]{Masanori~Ishikura}
\author[1]{Masayuki~Karasuyama\thanks{karasuyama@nitech.ac.jp}}
\affil[1]{Nagoya Institute of Technology}
\date{}

\newcommand{\biblstyle}{\bibliographystyle{apalike}}

\begin{document}
\maketitle
\begin{abstract}
Bayesian optimization often involves multiple objectives, constraints, and fidelity levels. We address the challenge of jointly selecting where and at which fidelity to evaluate to identify the highest-fidelity feasible Pareto frontier in this combined setting. From a unified information-theoretic perspective, we measure query utility by the information gain about this frontier, provided by an observation. Since this mutual information is intractable, we derive a variational lower bound using a mixture of under- and over-truncated approximations to the Pareto-consistent region. Multi-fidelity surrogate models propagate the information to arbitrary fidelities, yielding a cost-aware acquisition function without separate heuristics for fidelity selection or constraint handling. Experiments on synthetic, benchmark, and real-world problems demonstrate effectiveness across diverse objective, constraint, and fidelity settings.
\end{abstract}

\section{Introduction}
\label{sec:introduction}

Bayesian optimization (BO) is a powerful framework for optimizing expensive black-box functions. In many practical problems, however, several challenges arise simultaneously: multiple competing objectives must be optimized, constraints determine which designs are feasible, and evaluations are available at multiple fidelity levels with different costs and accuracies. These are inherently coupled when deciding both where and at which fidelity to evaluate. A useful query must account for its contribution to identifying favorable objective trade-offs, establishing feasibility, and reducing the need for expensive high-fidelity evaluations.

Several approaches have addressed subsets of these challenges, including constrained BO, multi-objective BO, and multi-fidelity BO. More recently, methods have been developed for the full constrained multi-fidelity multi-objective setting.
MFMO-SEGO \citep{Charayron2023-towards} and MF-CMOBO \citep{Lin2024-multi} address this setting using separate mechanisms for design, fidelity, and/or constraint handling.
MF-JESMOC \citep{FernandezSanchez2025-joint} instead provides a unified information-theoretic approach based on the joint entropy of the feasible Pareto set and frontier, but requires a more computationally involved inference procedure based on multi-fidelity deep Gaussian processes (MF-DGPs) and approximate joint conditioning on Pareto set and frontier.
These limitations motivate an alternative route toward a unified treatment of objectives, constraints, and fidelities with simpler inference.


We pursue such a formulation from the perspective of the feasible Pareto frontier at the target fidelity. Regardless of the fidelity at which an evaluation is performed, the ultimate goal is to identify this same frontier. We therefore measure the utility of a query at any input and fidelity by the information it provides about the target-fidelity feasible Pareto frontier, normalized by its evaluation cost. Under this view, multiple objectives, constraints, and fidelity selection are incorporated into a single information criterion rather than handled through separate acquisition mechanisms.

The resulting mutual information is, however, analytically intractable because conditioning on a Pareto frontier is difficult. To obtain a tractable criterion, we build on the variational information lower bound of PFEV \citep{ishikura2025pareto}. Our key construction expresses information about the frontier through Pareto-consistency events defined at the target fidelity, approximated by over- and under-truncated regions constructed from sampled Pareto frontiers. This event-based representation reduces the acquisition computation to Gaussian probabilities of these regions under the target-fidelity predictive distribution, before and after a prospective observation. For a query at any fidelity, the observation-conditioned distribution is obtained directly through standard cross-fidelity Gaussian conditioning. Thus, the unified information criterion can be evaluated using standard multi-fidelity GP inference.

Our main contributions are:
\begin{itemize}
 \setlength{\itemsep}{0pt}
 \item 
       We develop an information-theoretic approach to constrained multi-fidelity multi-objective BO centered on the feasible Pareto frontier at the target fidelity. The key idea is to represent frontier information through Pareto-consistency events defined at the target fidelity, allowing objectives, constraints, and fidelity selection to be handled within the same information criterion.
 \item 
       We derive a tractable variational lower bound based on over- and under-truncated approximations of these Pareto-consistency events by extending PFEV. The resulting acquisition computation reduces to Gaussian region probabilities before and after a prospective observation, where arbitrary-fidelity queries are connected to the target fidelity through standard cross-fidelity Gaussian conditioning.
 \item 
We evaluate the proposed method on GP-derived synthetic problems, benchmark functions, and real-world machine-learning hyperparameter optimization problems, demonstrating its effectiveness across different numbers of objectives, constraints, and input dimensions.
\end{itemize}

\section{Problem Setting}
\label{sec:problem}
We consider a family of multi-objective optimization problems with $L$ objective functions, $C$ constraint functions, and $M$ fidelity levels. 
Let $\cX\subseteq\RR^d$ be the design space.  
The $l$-th objective and the $c$-th constraint at fidelity $m \in [M] \coloneqq \{1,\ldots,M\}$ are denoted by $f_l^{(m)}:\cX\to\RR$ and $g_c^{(m)}:\cX\to\RR$, respectively. We write
$\*f^{(m)}_{\*x} \coloneqq (f_1^{(m)}(\*x),\ldots,f_L^{(m)}(\*x))^\top$,
$\*g^{(m)}_{\*x} \coloneqq (g_1^{(m)}(\*x),\ldots,g_C^{(m)}(\*x))^\top$
and
\begin{align*}
 \*h^{(m)}_{\*x}
 &\coloneqq
 \begin{bmatrix}
  \*f^{(m)}_{\*x}\\
  \*g^{(m)}_{\*x}
 \end{bmatrix}
 \in\RR^{L+C}.
\end{align*}
The highest fidelity $M$ is the target fidelity.  The optimization problem is
\begin{align*}
 \max_{\*x\in\cX} \ \*f^{(M)}_{\*x}, 
 \qquad
 \text{s.t. } \*g^{(M)}_{\*x}\geq \*z,
\end{align*}
where $\*z=(z_1,\ldots,z_C)^\top$ is the vector of constraint thresholds and vector inequalities are interpreted elementwise.  
We refer to this constrained multi-fidelity multi-objective optimization problem as a C-MF-MO problem.
%
For two objective vectors $\*a,\*b\in\RR^L$, we say that $\*a$ dominates $\*b$, written as $\*a\succ\*b$, if $a_l\geq b_l$ for every $l$ and the inequality is strict for at least one objective.  The feasible set at the target fidelity is
$\cX_{\mathrm{feas}}
 \coloneqq \{\*x\in\cX\mid \*g^{(M)}_{\*x}\geq\*z \}$,
and the target Pareto frontier is
\begin{align*}
 \cF^\ast\coloneqq
 \left\{\*f^{(M)}_{\*x}\;\middle|\;
 \begin{aligned}
 &\*x\in\cX_{\mathrm{feas}},\\
 &\nexists\,\*x'\in\cX_{\mathrm{feas}}:\
   \*f^{(M)}_{\*x'}\succ\*f^{(M)}_{\*x}
 \end{aligned}
 \right\}.
\end{align*}
Our goal is to approximate the highest-fidelity feasible Pareto frontier $\cF^\ast$ under a finite evaluation budget.  Accordingly, an observation at any fidelity is valued by the information it provides about $\cF^\ast$.

\subsection{Multi-fidelity GP and observation model}
\label{sec:gp_noise}
Each one of objective and constraint functions is modeled by an independent multi-fidelity Gaussian process (GP), resulting in $L+C$ independent $M$-output vector-valued GPs \citep{alvarez2012kernels}.
%
We represent fidelity as an additional GP input, so different fidelities of the same scalar output are statistically coupled through the GP covariance.
For example, each objective function is modeled as
\begin{align*}
 f_{\ell}^{(m)}(\*x) \sim \cG\cP(0,k((\*x,m),(\*x^\prm,m^\prm))),
\end{align*}
where $k$ is a joint kernel function for an input $\*x$ and a fidelity $m$, and the constraint function is defined by the same way.
%
%
Conditioning on the current dataset $\cD$ is omitted below when no confusion arises. At a query $(\*x,m)$, we observe a noisy version of the latent output:
\begin{align*}
 \*r^{(m)}_{\*x}
 \coloneqq
 \*h^{(m)}_{\*x}+\*\epsilon^{(m)}_{\*x},
 \qquad
 \*\epsilon^{(m)}_{\*x}
 \sim \cN\rbr{\*0,\*\Sigma_{\epsilon}^{(m)}}.
\end{align*}
We assume that the observation noises are independent across scalar outputs and independent of the latent functions.  
Accordingly, $\*\Sigma_{\epsilon}^{(m)}$ is diagonal.  
%
%
Let $c_m>0$ denote the evaluation cost at fidelity $m$.

\section{A Unified Information-Theoretic Acquisition Function}
\label{sec:method}

Our basic idea is that a query is useful if its observation reduces uncertainty about the feasible target-fidelity Pareto frontier.  
We therefore consider the following mutual information (MI) between an observation 
$\*r^{(m)}_{\*x}$
and 
the Pareto-frontier $\cF^\ast$: 
\begin{align}
 \mr{MI}\rbr{\*r^{(m)}_{\*x}; \cF^\ast}.
 \label{eq:MI}
\end{align}
%
An advantage of this criterion is that we do not require any external mechanisms for incorporating each of multiple objectives, constraints, and fidelity, separately, because all of them are naturally incorporated into dependency structure in MI.
%
In this sense, this MI can be seen as a unified criterion for our C-MF-MO problem, nevertheless it is a highly complicated problem setting for BO.
%
%

\subsection{Variational information lower bound}
\label{sec:vlb_noisy}

Since direct evaluation of MI \eq{eq:MI} is highly complicated, we instead employ a variational approach.
Let $q(\*r^{(m)}_{\*x}\mid\cF^\ast)$ be a variational conditional density whose support contains that of $p(\*r^{(m)}_{\*x}\mid\cF^\ast)$.  
The mutual information admits the following lower bound:
\begin{align}
 &\mr{MI}\rbr{\*r^{(m)}_{\*x};\cF^\ast}
 \geq
 \EE_{\cF^\ast,\*r^{(m)}_{\*x}}
 \sbr{
   \log
   \frac{q(\*r^{(m)}_{\*x}\mid\cF^\ast)}
        {p(\*r^{(m)}_{\*x})}
 }. 
 \label{eq:noisy_vlb}
\end{align}
The derivation is in Appendix~\ref{app:variational-lower-bound}, which follows the lower-bound construction introduced for constrained BO by \citet{Takeno2022-Sequential} and used by PFEV for Pareto-frontier information \citep{ishikura2025pareto}. 
On the other hand, any application to the C-MF-MO problem has not been explored.
We call our method CMF-PFEV, because it exploits the idea of mixture-based variational density of PFEV.

\subsection{Local Pareto conditioning and finite-frontier approximation}
\label{sec:truncation_events}

The variational conditional density 
$q(\*r^{(m)}_{\*x} \mid \cF^\ast)$
is a surrogate of 
$p(\*r^{(m)}_{\*x} \mid \cF^\ast)$, 
whose analytical representation is not tractable because GPs conditioned on $\cF^\ast$ has not been revealed even for the unconstrained single-fidelity problem \citep{Suzuki2020-multi}.
%
We replace conditioning on the entire frontier by a `Pareto-consistency' condition on the target-fidelity output at the candidate input $\*x$. 
We refer to Pareto-consistency as the fact that given $\cF^\ast$, the output $\*h^{(M)}_{\*x}$ should be in the following admissible region $\cA^{\cF^\ast}$ defined as
\begin{align*}
 \cA^{\cF^\ast}
 \coloneqq
 \left\{\*h^{(M)}\;\middle|\;
 \begin{aligned}
 &(\*g^{(M)}\not\geq\*z)\;\lor\\[-1pt]
 &\rbr{(\*f^{(M)}\preceq\cF^\ast)\land(\*g^{(M)}\geq\*z)}
 \end{aligned}
\right\},
\end{align*}
where
$\*h^{(M)}=((\*f^{(M)})^\top,(\*g^{(M)})^\top)^\top$ 
is a target-fidelity output 
and 
$\*f^{(M)} \preceq \cF^\ast$ means that $\*f^{(M)}$ is dominated by or equal to at least one element of $\cF^\ast$.
If $\*h^{(M)}_{\*x}$ exists outside of $\cA^{\cF^\ast}$, it contradicts with the fact that $\cF^\ast$ is Pareto optimal.
As shown in Fig.~\ref{fig:admissible_regions}~(a),
$\cA^{\cF^\ast}$ 
contains outputs that are either infeasible or have objective values dominated by or equal to some frontier point.  
We define the variational distribution by imposing Pareto-consitency condition $\*h^{(M)}_{\*x}\in\cA^{\cF^\ast}$ as
\begin{align}
 q\rbr{\*r^{(m)}_{\*x}\mid\cF^\ast}
 & \coloneqq
 p\rbr{\*r^{(m)}_{\*x}\mid
 \*h^{(M)}_{\*x}\in\cA^{\cF^\ast}}
 \label{eq:local_pareto_conditional}
\end{align}
%
%
Thus, \eq{eq:local_pareto_conditional} represents that a feasible target-fidelity output cannot exist beyond the frontier $\cF^\ast$.

Our variational distribution \eq{eq:local_pareto_conditional} can be seen as a local approximation: the event at $\*x$ retains a necessary condition implied by $\cF^\ast$, but does not impose all the conditions on the latent functions across the design space. 
Note that this event concerns the latent output at fidelity $M$, even when the observation is made at a lower fidelity $m$.

In a continuous design space, the frontier $\cF^\ast$ may contain infinitely many points.
In the actual computation, we need to approximate it by a set of finite sampled points in $\cF^\ast$.
We represent a sampled frontier by a finite set $\widehat{\cF}^\ast \subseteq \cF^\ast$ of $N_{\mathrm{PF}}$ points.
%
%
%
Following PFEV, we use it to construct two event-conditioned densities and combine them into the variational distribution, with $\lambda\in(0,1]$,
\begin{align}
 q_\lambda\rbr{\*r^{(m)}_{\*x}\mid\widehat{\cF}^\ast}
 &=\lambda q_U\rbr{\*r^{(m)}_{\*x}\mid\widehat{\cF}^\ast}
 \notag\\
 &\quad 
 +(1-\lambda)q_O\rbr{\*r^{(m)}_{\*x}\mid\widehat{\cF}^\ast}.
 \label{eq:cmf_mixture}
\end{align}
For $T\in\{O,U\}$, the component density is
\begin{align}
 & q_T\rbr{\*r^{(m)}_{\*x}\mid \widehat{\cF}^\ast}
 \coloneqq 
 p\rbr{\*r^{(m)}_{\*x}\mid \*h^{(M)}_{\*x}\in\cA_T^{\widehat{\cF}^\ast}}
 \notag \\ 
 & \qquad =
 \frac{
 \Pr\rbr{
 \*h^{(M)}_{\*x}\in \cA_T^{\widehat{\cF}^\ast}
 \mid
 \*r^{(m)}_{\*x}
 }
 p\rbr{
 \*r^{(m)}_{\*x}
 }}{
 \Pr\rbr{
 \*h^{(M)}_{\*x}\in \cA_T^{\widehat{\cF}^\ast}
 }
 }.
 \label{eq:q_T}
\end{align}
In the last line, we apply Bayes theorem for later computations. 
A key difference between $q_U$ and $q_O$ is in the two regions $\cA_T^{\widehat{\cF}^\ast}$ defined as follows.  
%
Let
\begin{align*}
 \mathcal{R}_O(\widehat{\cF}^\ast)
 &\coloneqq
 \cbr{\*f^{(M)}\in\RR^L\mid
 \exists\,\*p\in\widehat{\cF}^\ast:\ \*f^{(M)}\preceq\*p},\\
 \mathcal{R}_U(\widehat{\cF}^\ast)
 &\coloneqq
 \cbr{\*f^{(M)}\in\RR^L\mid
 \nexists\,\*p\in\widehat{\cF}^\ast:\ \*f^{(M)}\succ\*p}.
\end{align*}
The former contains vectors dominated by or equal to a sampled frontier point; the latter excludes vectors that strictly dominate any sampled frontier point. Here strict dominance means being no worse in every objective and strictly better in at least one, as defined in Section~\ref{sec:problem}; equality with a sampled frontier point is therefore retained in $\mathcal{R}_U$.  Including infeasible outputs in both cases gives
\begin{align*}
 \cA_T^{\widehat{\cF}^\ast}
 \coloneqq
 \left\{\*h^{(M)}\;\middle|\;
 \begin{aligned}
 &(\*g^{(M)}\not\geq\*z)\;\lor\\[-1pt]
 &\rbr{(\*f^{(M)}\in\mathcal{R}_T)\land(\*g^{(M)}\geq\*z)}
 \end{aligned}
\right\},
\end{align*}
where the dependence of $\mathcal{R}_T$ on $\widehat{\cF}^\ast$ is omitted.
%
The regions satisfy
$\cA_O^{\widehat{\cF}^\ast}
 \subseteq\cA^{\cF^\ast}
 \subseteq \cA_U^{\widehat{\cF}^\ast}$.
Accordingly, $O$ and $U$ denote over- and under-truncation, respectively, by which we expect balancing these two opposite approximations can provide a better variational approximation.
%
Figure~\ref{fig:admissible_regions}~(b) and (c) illustrate
$\cA_O^{\widehat{\cF}^\ast}$
and 
$\cA_U^{\widehat{\cF}^\ast}$,
respectively.
%
%
We retain $\lambda>0$ so that the under-truncated component also covers the required support; a purely over-truncated density may otherwise violate the support condition in \eqref{eq:noisy_vlb}. 
\iftwocol\begin{figure*}[t]\else\begin{figure}[t]\fi
\centering
\includegraphics[width=0.85\textwidth]{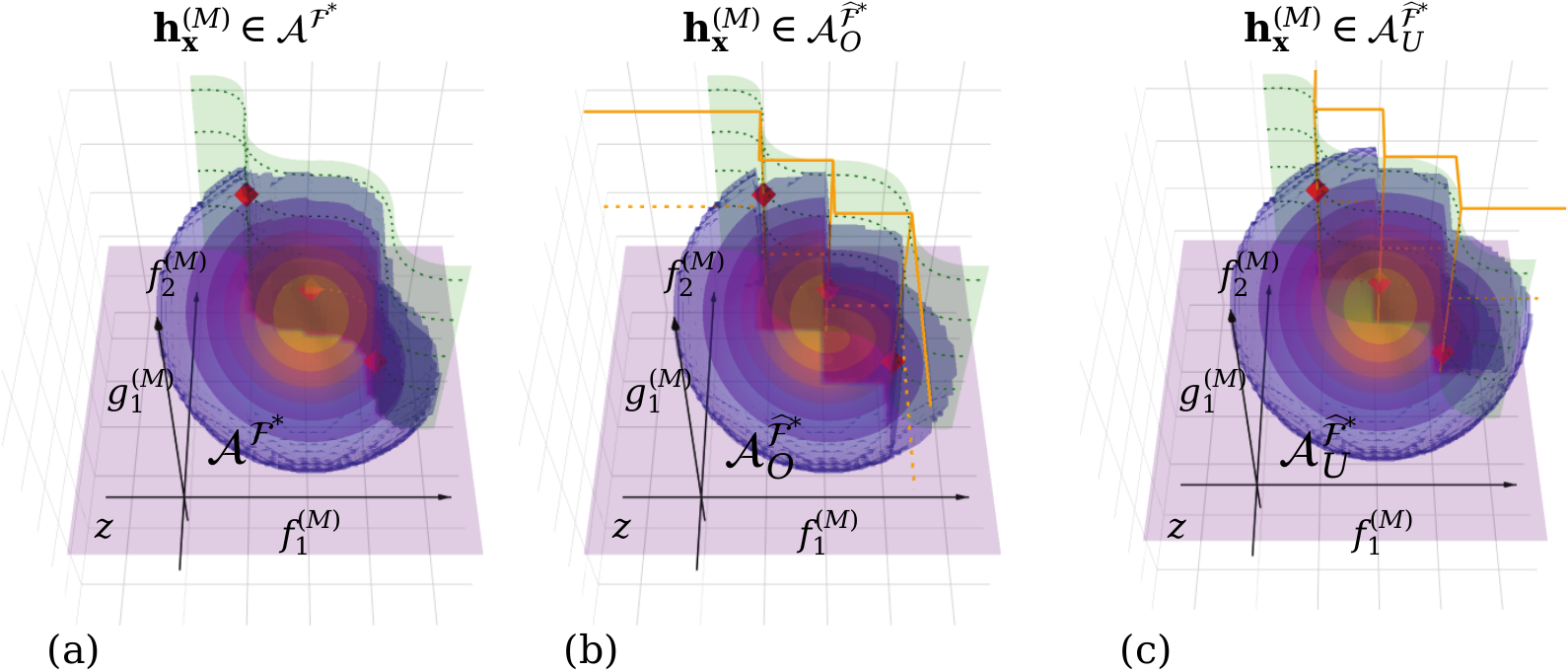}
\caption{
Admissible target-fidelity regions for two objectives and one constraint.
The $Z$-axis represents the constraint $g_1^{(M)}$, for which below the threshold $z$ (purple level plane) is infeasible. 
The $XY$-plane corresponds to two objectives $f_1^{(M)}$ and $f_2^{(M)}$, on which the Pareto frontier $\cF^\ast$ is defined, depicted as the green wavy plane. 
(a)
The distribution of $\*h_{\*x}^{(M)}$ (the sphere) is truncated by $\cF^\ast$ and $g_1^{(M)}(\*x) = z$. 
(b) 
Based on sampled $\widehat{\cF}^\ast$ depicted as the red diamonds, the approximate truncated region is constructed by the over-truncation (the yellow zigzag plane). 
(c)
Same as (b), the approximate truncated region is constructed by the under-truncation (the yellow zigzag plane). 
The truncations in (b) and (c) are two extreme approximations of the unknown truncation in (a). 
Equation \eq{eq:cmf_mixture} uses a mixture of their truncated distributions as the variational distribution.
%
}
\label{fig:admissible_regions}
\iftwocol\end{figure*}\else\end{figure}\fi

\subsection{From the variational mixture to event probabilities}
\label{sec:cmf_pfev}
Substituting the mixture in \eqref{eq:cmf_mixture} into \eqref{eq:noisy_vlb} yields the lower bound
\begin{align}
 &\mr{MI}\rbr{\*r^{(m)}_{\*x};\cF^\ast}
 \geq
 \cL_{\mathrm{CMF}}(\*x,m,\lambda)
 \notag
 \\
 & 
 \coloneqq
 \EE_{\cF^\ast,\*r^{(m)}_{\*x}}
 \left[\log\left\{
 \begin{aligned}
 &\lambda\frac{q_U(\*r^{(m)}_{\*x}\mid\widehat{\cF}^\ast)}{p(\*r^{(m)}_{\*x})}\\
 &{}+(1-\lambda)\frac{q_O(\*r^{(m)}_{\*x}\mid\widehat{\cF}^\ast)}{p(\*r^{(m)}_{\*x})}
 \end{aligned}
 \right\}\right].
 \label{eq:cmf_density_lb}
\end{align}
The expectation is taken under the joint posterior of the frontier $\cF^\ast$ and the prospective observation $\*r^{(m)}_{\*x}$. 
Inside the expectation, only the finite frontier $\widehat{\cF}^\ast$ is used, rather than the original $\cF^\ast$ that is not necessarily available.

We need to evaluate the density ratios inside the logarithm.
Both the numerator and denominator can be expressed using the probability of the same target-fidelity region, before and after a prospective observation is given. 
We define
\begin{align}
 Z_T(\*x;\widehat{\cF}^\ast)
 &
 \coloneqq
 \Pr\rbr{ \*h^{(M)}_{\*x}\in\cA_T^{\widehat{\cF}^\ast}},
 \label{eq:Z_T}
 \\
 Z_T(\*x;\widehat{\cF}^\ast\mid\*r^{(m)}_{\*x})
 &
 \coloneqq
 \Pr\rbr{\*h^{(M)}_{\*x}\in\cA_T^{\widehat{\cF}^\ast}
 \mid\*r^{(m)}_{\*x}}.
 \label{eq:Z_T-given-r}
\end{align}
The second quantity is the observation-conditioned version of the first. Below, we abbreviate these quantities as $Z_T$ and $Z_T(\cdot\mid\*r^{(m)}_{\*x})$, respectively, where the dot represents the omitted arguments $\*x$ and $\widehat{\cF}^\ast$. 
%
Combining \eq{eq:Z_T}, \eq{eq:Z_T-given-r}, and \eq{eq:q_T}, we obtain
\begin{align}
 {q_T(\*r^{(m)}_{\*x}\mid\widehat{\cF}^\ast)} / 
 {p(\*r^{(m)}_{\*x})}
 =
 {Z_T(\cdot\mid\*r^{(m)}_{\*x})} /
 {Z_T}.
 \label{eq:bayes_event_ratio}
\end{align}
The lower bound to be evaluated is therefore
\begin{align}
 &\cL_{\mathrm{CMF}}(\*x,m,\lambda)
 \notag\\
 & \ =
 \EE_{\cF^\ast,\*r^{(m)}_{\*x}}
 \left[\log\left\{
 \begin{aligned}
 &\lambda\frac{Z_U(\cdot\mid\*r^{(m)}_{\*x})}{Z_U}\\
 &{}+(1-\lambda)\frac{Z_O(\cdot\mid\*r^{(m)}_{\*x})}{Z_O}
 \end{aligned}
 \right\}\right].
 \label{eq:cmf_pfev_lb}
\end{align}
Thus, we need the probability of the same Pareto-consistency event 
$\*h^{(M)}_{\*x}\in\cA_T^{\widehat{\cF}^\ast}$ 
before and after a prospective observation, i.e., 
$Z_T$
and
$Z_T(\cdot\mid\*r^{(m)}_{\*x})$.
A low-fidelity query can be regarded as useful insofar as its observation changes this target-fidelity event probability.
%
To determine $\lambda$, we maximize \eq{eq:cmf_pfev_lb} over $\lambda$.
This is a special case of the well-known variational lower bound maximization, justified as the minimization the KL divergence between the variational distribution $q_\lambda$ and the true distribution (See Appendix~\ref{app:variational-lower-bound}).
%
%

\subsection{Cross-fidelity propagation and region probabilities}
\label{sec:cmf_exact_prob}
To evaluate the denominator $Z_T$, we use the target-fidelity marginal 
$\*h_{\*x}^{(M)} \sim \cN(\*\mu_M,\*\Sigma_{MM})$, 
where $\*\mu_M$ and $\*\Sigma_{MM}$ are the posterior mean vector and variance matrix of $\*h_{\*x}^{(M)}$, respectively (since $\*h_{\*x}^{(M)}$ is now modeled by independent GPs, $\*\Sigma_{MM}$ is a diagonal matrix). 
To evaluate the numerator $Z_T(\cdot\mid\*r^{(m)}_{\*x})$, we first condition this distribution on the prospective observation.  
Let the joint posterior be
\begin{align*}
 \begin{bmatrix}
  \*h^{(M)}_{\*x}\\
  \*r^{(m)}_{\*x}
 \end{bmatrix}
 \sim
 \cN\left(
 \begin{bmatrix}
  \*\mu_M\\
  \*\mu_m
 \end{bmatrix},
 \begin{bmatrix}
  \*\Sigma_{MM} & \*\Sigma_{Mm}\\
  \*\Sigma_{mM} &
  \*\Sigma_{mm}+\*\Sigma_{\epsilon}^{(m)}
 \end{bmatrix}
 \right).
\end{align*}
Hence, after a hypothetical observation $\*r^{(m)}_{\*x}$,
\begin{align*}
 \*h^{(M)}_{\*x}\mid \*r^{(m)}_{\*x}
 \sim \cN\rbr{\*\mu_{M\mid r},\*\Sigma_{M\mid r}},
\end{align*}
where
\begin{align*}
 \*\mu_{M\mid r}
 &=\*\mu_M
 +\*\Sigma_{Mm}
 \rbr{\*\Sigma_{mm}+\*\Sigma_{\epsilon}^{(m)}}^{-1}
 \rbr{\*r^{(m)}_{\*x}-\*\mu_m},\\
 \*\Sigma_{M\mid r}
 &=\*\Sigma_{MM}
 -\*\Sigma_{Mm}
 \rbr{\*\Sigma_{mm}+\*\Sigma_{\epsilon}^{(m)}}^{-1}
 \*\Sigma_{mM}.
\end{align*}
We write $\mu_{f,l\mid r}$ and $\mu_{g,c\mid r}$ for the objective and constraint components of $\*\mu_{M\mid r}$, and $\sigma_{f,l\mid r}$ and $\sigma_{g,c\mid r}$ for the square roots of the corresponding diagonal entries of $\*\Sigma_{M\mid r}$. 

This equation makes the role of fidelity explicit: a query is informative about the target frontier only insofar as the cross-fidelity covariance allows its observation to change posterior beliefs about the target-fidelity outputs. Let
\begin{align*}
 \rho_{\mathrm{feas}}(\*r^{(m)}_{\*x})
 &\coloneqq
 \Pr\rbr{\*g^{(M)}_{\*x}\geq\*z\mid\*r^{(m)}_{\*x}},\\
 \rho_O(\*r^{(m)}_{\*x})
 &\coloneqq
 \Pr\rbr{\*f^{(M)}_{\*x}\in\mathcal{R}_O
 \mid\*r^{(m)}_{\*x}},\\
 \rho_U(\*r^{(m)}_{\*x})
 &\coloneqq
 \Pr\rbr{\*f^{(M)}_{\*x}\in\mathcal{R}_U
 \mid\*r^{(m)}_{\*x}}.
\end{align*}
Independence of the scalar-output GPs, together with the independent-noise assumption, gives
\begin{align*}
 Z_O(\cdot\mid\*r^{(m)}_{\*x})
 &= ( 1-\rho_{\mathrm{feas}}(\*r^{(m)}_{\*x}) )
 +\rho_{\mathrm{feas}}(\*r^{(m)}_{\*x})
 \rho_O(\*r^{(m)}_{\*x}),
 \\
 Z_U(\cdot\mid\*r^{(m)}_{\*x})
 &= ( 1-\rho_{\mathrm{feas}}(\*r^{(m)}_{\*x}) )
 +\rho_{\mathrm{feas}}(\*r^{(m)}_{\*x})
 \rho_U(\*r^{(m)}_{\*x}).
\end{align*}
We see that $\rho_{\mathrm{feas}}$ is written as
\begin{align}
 \rho_{\mathrm{feas}}(\*r^{(m)}_{\*x})
 =\prod_{c=1}^{C}
 \left[
  1-\Phi\left(
  \frac{z_c-\mu_{g,c\mid r}}{\sigma_{g,c\mid r}}
  \right)
 \right],
 \label{eq:pof_conditional}
\end{align}
where $\Phi$ is the standard normal CDF.
For CMF-PFEV, the objective-space probabilities are evaluated by partitioning the region dominated by the finite frontier into mutually disjoint hyper-rectangles, as in Pareto-frontier entropy search \citep{Suzuki2020-multi,ishikura2025pareto}. Suppose that
\begin{align*}
 \mathcal{R}_O=\bigcup_{i=1}^{N_O}\cC_i,
 \qquad
 \cC_i=\prod_{l=1}^{L}(a_{i,l},b_{i,l}],
\end{align*}
where $N_O$ denotes the number of disjoint cells, and $a_{i,l}$ and $b_{i,l}$ are the lower and upper bounds of $\cC_i$ along objective dimension $l$, respectively. 
Then, we see
\begin{align*}
 &\rho_O(\*r^{(m)}_{\*x})
 \! = \!
 \sum_{i=1}^{N_O}\prod_{l=1}^{L}
 \!
 \left[ \!
 \Phi \! \left(\frac{b_{i,l}-\mu_{f,l\mid r}}{\sigma_{f,l\mid r}}\right)
 \! - \! 
 \Phi \! \left(\frac{a_{i,l}-\mu_{f,l\mid r}}{\sigma_{f,l\mid r}}\right)
 \! \right].
\end{align*}
The complement probability $1-\rho_U$, corresponding to the region of objective vectors that strictly dominate at least one sampled frontier point, is obtained analogously; the same dominated-region routine can be applied after sign-flipping the objective vectors. 
The unconditional normalizers $Z_T$ are obtained by the same region calculations under the unconditional target-fidelity predictive distribution.

\subsection{Monte Carlo approximation and cost-aware query selection}
\label{sec:outer_mc}
The expectation over the random frontier and prospective noisy observation in \eqref{eq:cmf_pfev_lb} is approximated by nested Monte Carlo sampling. For each $k=1,\ldots,N_{\mathrm{path}}$, we draw a posterior function path and solve the constrained multi-objective problem at fidelity $M$ to obtain the finite frontier $\widetilde{\cF}_k^\ast$ containing $N_{\mathrm{PF}}$ points.  The latent query output $\widetilde{\*h}_{\*x,k}^{(m)}$ is evaluated on the same function path, thereby preserving its posterior dependence with the sampled target frontier.  For each path, we then draw $N_{\epsilon}$ independent observation-noise samples and set
\begin{align*}
 \widetilde{\*r}_{\*x,k,j}^{(m)}
 &=\widetilde{\*h}_{\*x,k}^{(m)}
   +\widetilde{\*\epsilon}_{k,j}^{(m)}
\end{align*}
for
$\widetilde{\*\epsilon}_{k,j}^{(m)}
 \overset{\mathrm{i.i.d.}}{\sim}
 \cN(\*0,\*\Sigma_{\epsilon}^{(m)}),
  j=1,\ldots,N_{\epsilon}$,
where the noise samples are independent of the posterior function paths.  The lower bound is approximated as
\begin{align}
 &\widehat{\cL}_{\mathrm{CMF}}(\*x,m,\lambda)
 =\frac{1}{N_{\mathrm{path}}N_{\epsilon}}
 \sum_{k=1}^{N_{\mathrm{path}}}
 \sum_{j=1}^{N_{\epsilon}}
 \label{eq:outer_mc_lb} \\
 & \log \Biggl\{ 
 \lambda\frac{Z_{U,k}(\cdot\mid\widetilde{\*r}_{\*x,k,j}^{(m)})}{Z_{U,k}}
 +(1-\lambda)\frac{Z_{O,k}(\cdot\mid\widetilde{\*r}_{\*x,k,j}^{(m)})}{Z_{O,k}}
 \Biggr\}
 \notag
\end{align}
Both the posterior function paths and prospective observation noise are sampled by ordinary Monte Carlo.  The Gaussian base-noise samples are held fixed during acquisition optimization to provide common random numbers across candidate points.  
Here $Z_{T,k}$ and $Z_{T,k}(\cdot\mid\widetilde{\*r}_{\*x,k,j}^{(m)})$ denote the unconditional and observation-conditioned region probabilities evaluated using the finite frontier $\widetilde{\cF}_k^\ast$ from path $k$.  
Finally, let $c_m$ be the cost of the coupled query at fidelity $m$. 
We evaluate information gain per unit cost at a given mixture weight,
\begin{align*}
 \alpha(\*x,m,\lambda)
 &=\frac{1}{c_m}
 \widehat{\cL}_{\mathrm{CMF}}(\*x,m,\lambda).
\end{align*}
For each discrete fidelity $m\in\{1,\ldots,M\}$, we jointly optimize the candidate input and mixture weight:
\begin{align*}
 (\*x_m,\lambda_m)
 &\in\argmax_{\substack{\*x\in\cX\\\lambda\in(0,1]}}
 \alpha(\*x,m,\lambda).
\end{align*}
In the continuous design setting, this is a joint continuous optimization over $(\*x,\lambda)$ with $m$ held fixed. 
We then select the fidelity by 
$\argmax_{m \in \{1,\ldots,M\}} \alpha(\*x_m,m,\lambda_m)$, 
which indicates the most cost-effective fidelity in a sense of the approximate MI.


The complete Bayesian optimization loop is summarized in Algorithm~\ref{alg:cmf_pfev} in Appendix~\ref{app:procedure}. 

\begin{figure*}[t] 

 \centering

 \subfloat[][$(L,C,d) = (2,1,2)$]{
 \ig{.3}{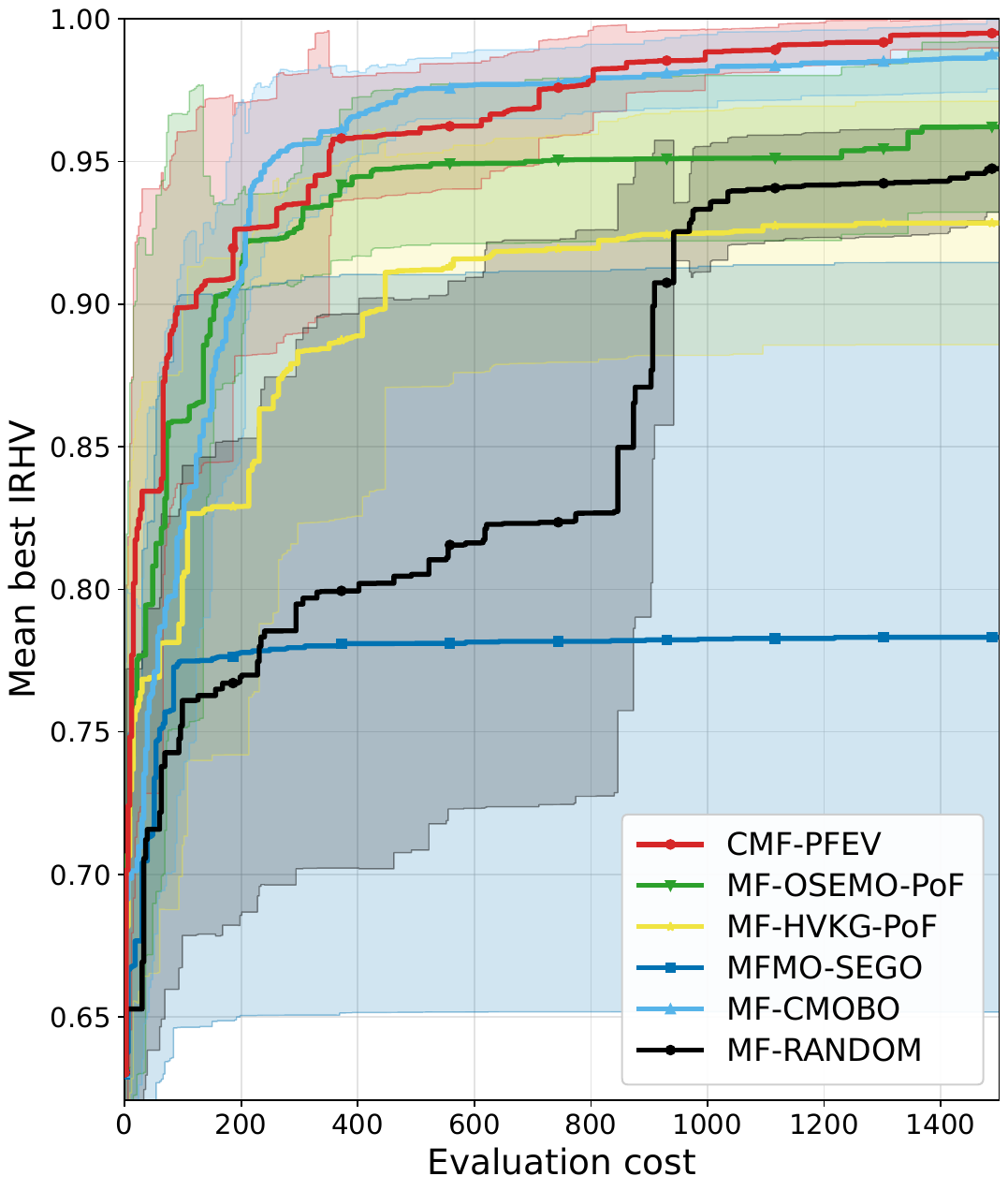}}
 \subfloat[][$(L,C,d) = (2,1,5)$]{
 \ig{.3}{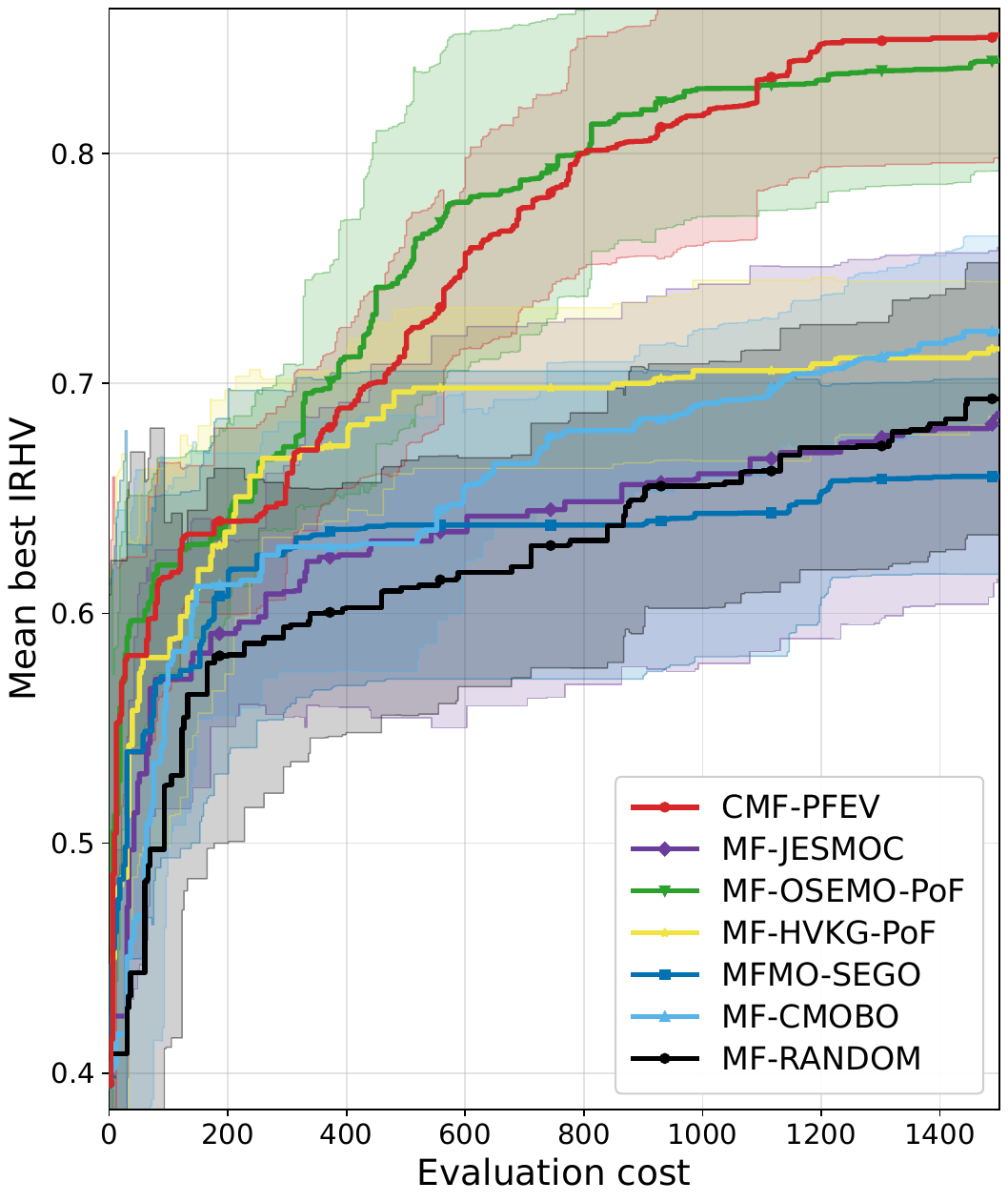}}
 \subfloat[][$(L,C,d) = (3,1,5)$]{
 \ig{.3}{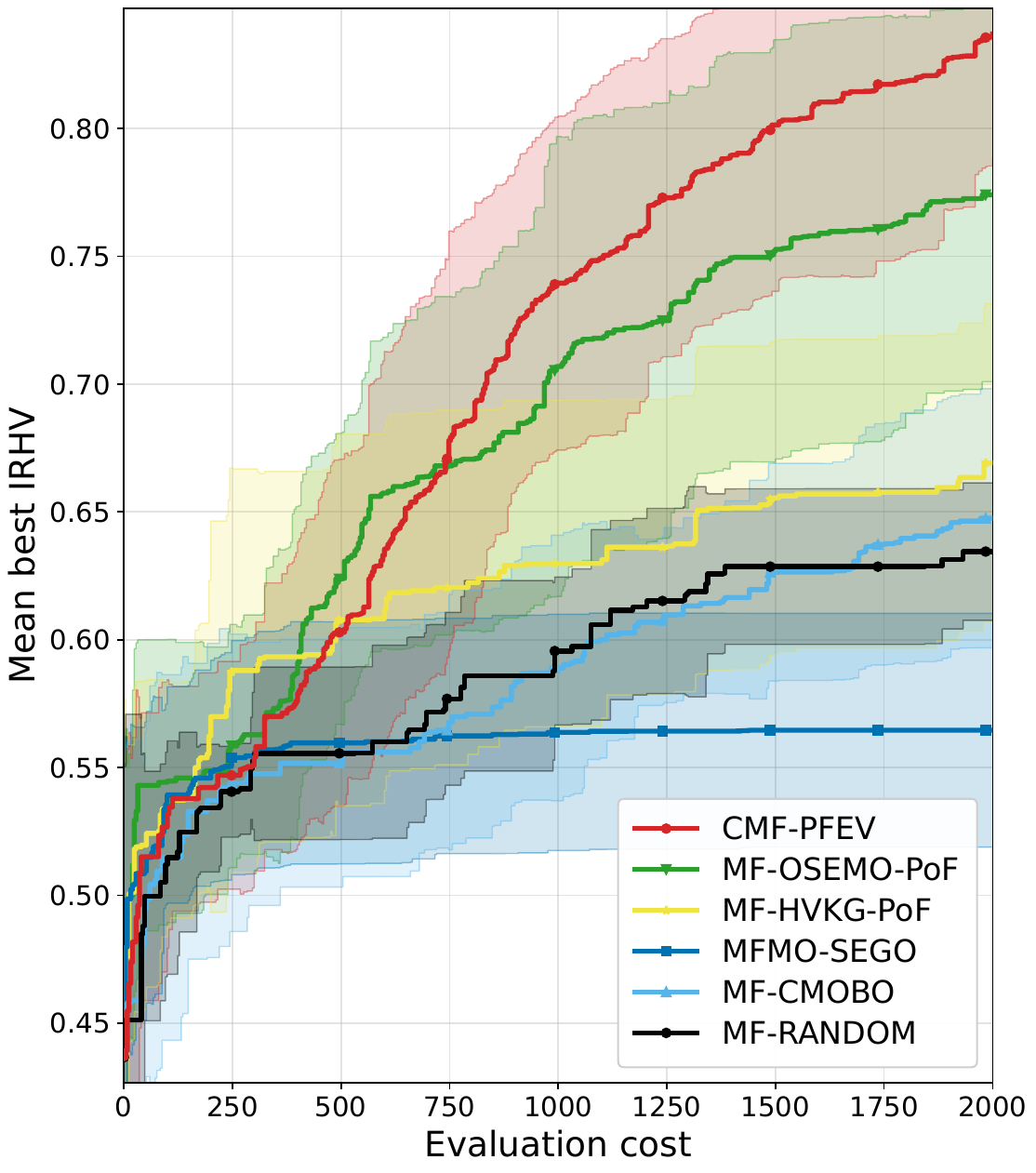}}

 \subfloat[][$(L,C,d) = (2,2,2)$]{
 \ig{.3}{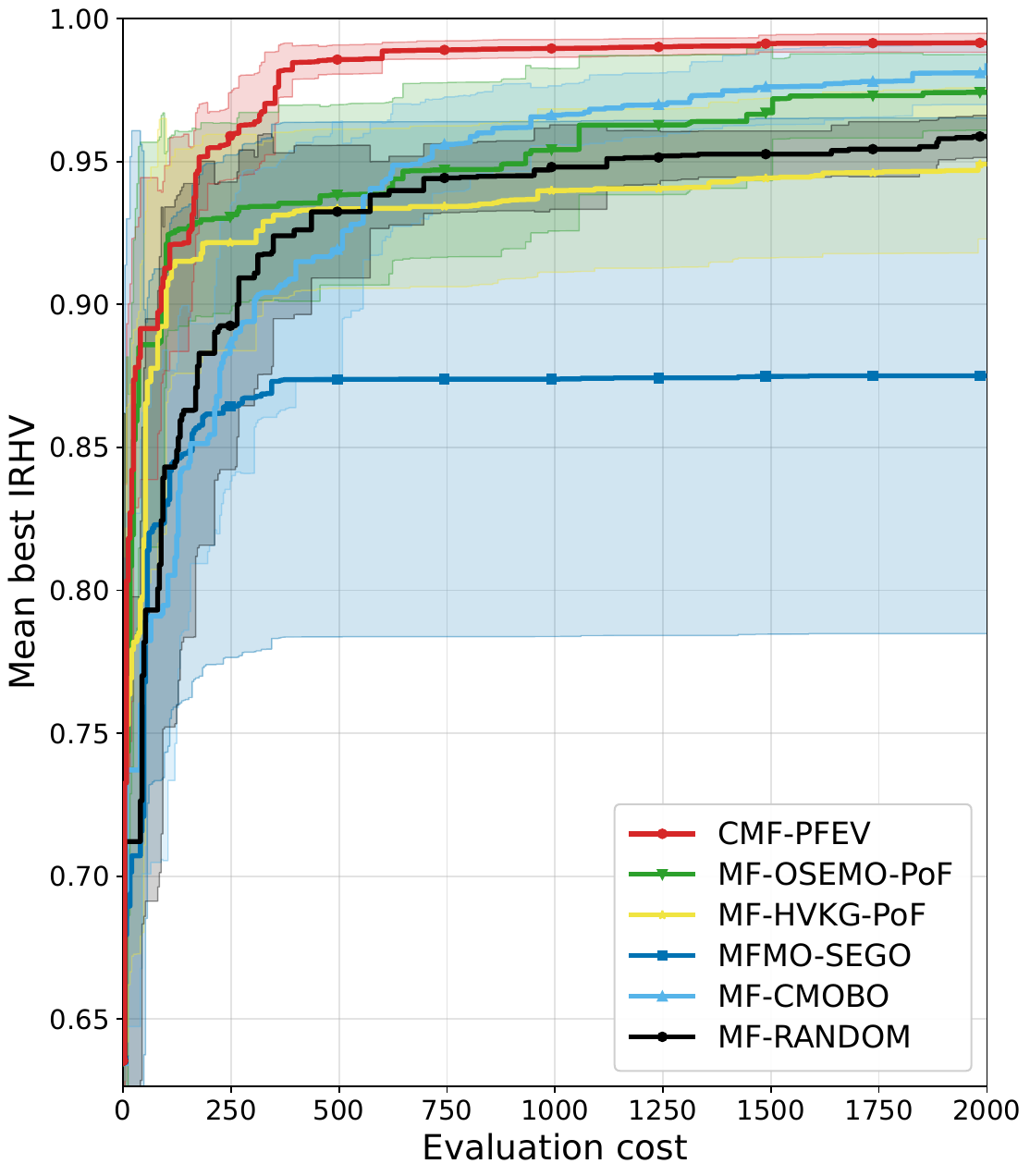}}
 \subfloat[][$(L,C,d) = (2,2,5)$]{
 \ig{.3}{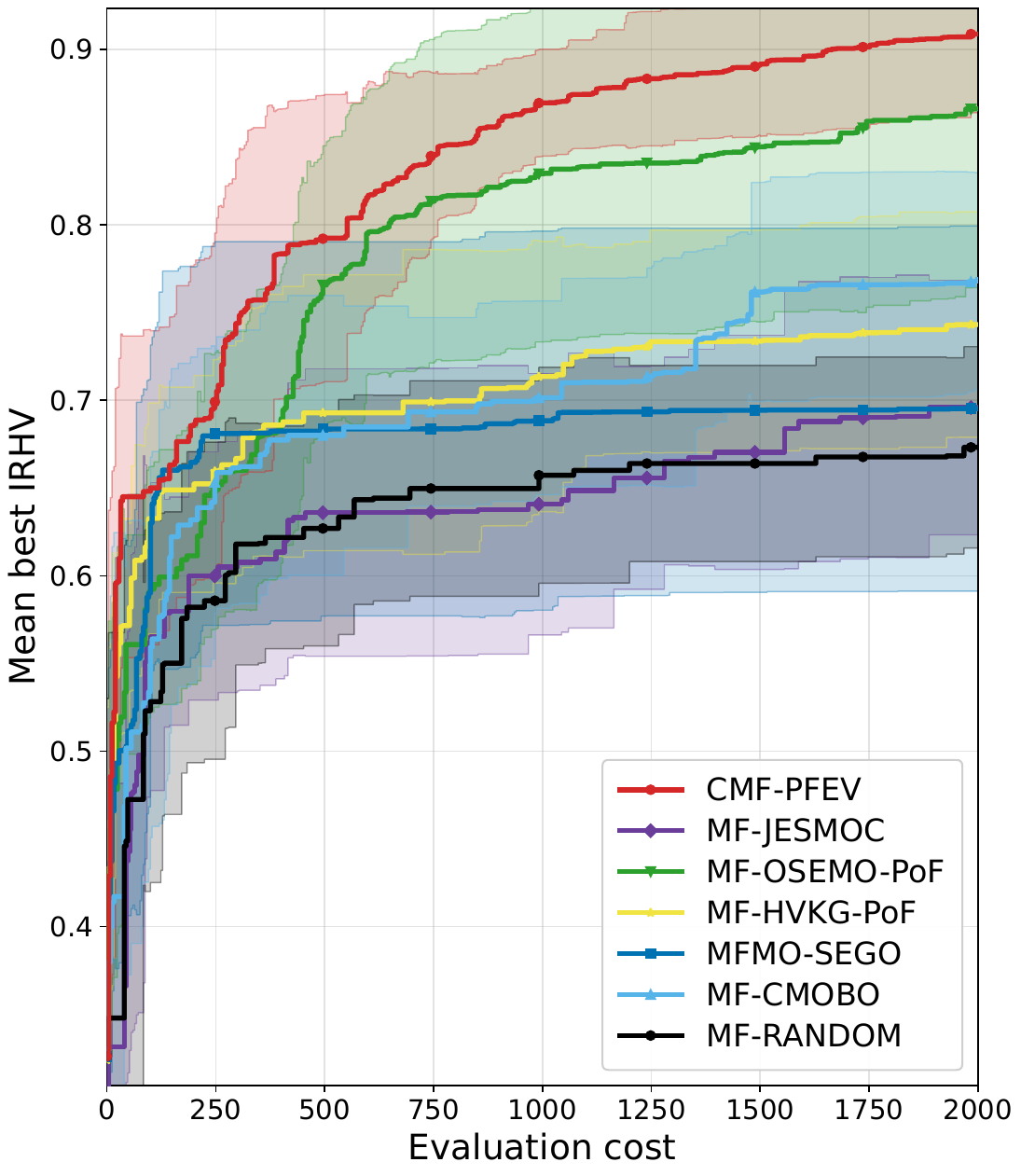}}
 \subfloat[][$(L,C,d) = (3,2,5)$]{
 \ig{.3}{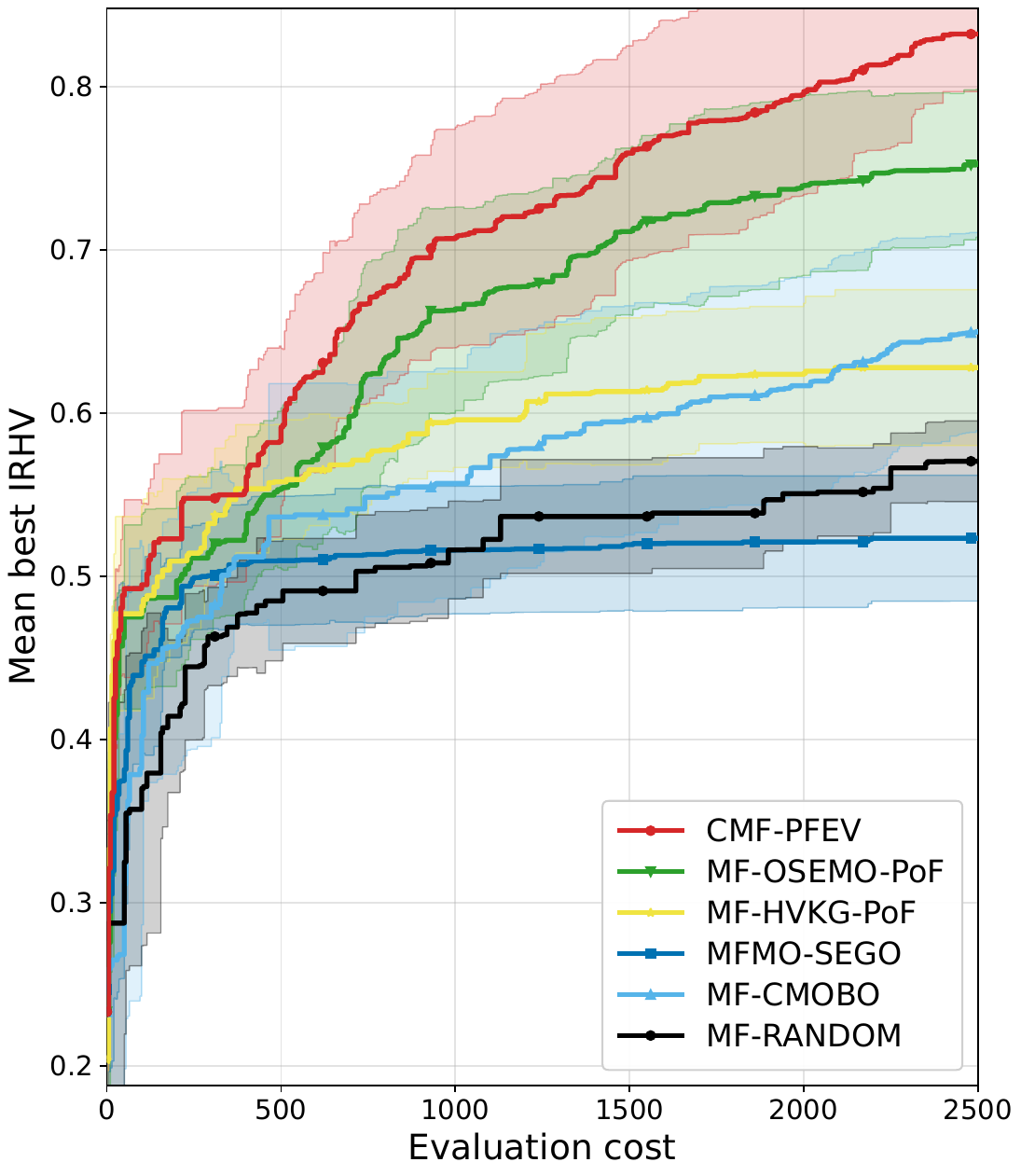}}

 \caption{Results on GP-derived synthetic functions (10 runs average and standard error).}
 \label{fig:result-gp-data}

\end{figure*}

\section{Related Work}
\label{sec:related_work}

{\bf Multi-objective BO.}
Multi-objective BO (MOBO) includes scalarization-based approaches such as ParEGO \citep{Knowles2006-parego}, confidence-bound approaches such as PAL \citep{Zuluaga2013-active}, and hypervolume-based criteria such as EHVI and its Monte Carlo extensions \citep{daulton20differentiable}.
Information-based methods instead reduce uncertainty about Pareto-optimal quantities: PESMO targets the Pareto set \citep{Hernandez-lobatoa2016-Predictive}, MESMO, PFES, and PF2ES target the Pareto frontier \citep{Belakaria2019-maxvalue,Suzuki2020-multi,qing2023pf}, and JES jointly targets the Pareto set and frontier \citep{Tu2022-joint}. 
PFEV constructs a variational information lower bound using over- and under-truncated approximations to the Pareto-consistent region (Ishikura and Karasuyama, 2025), on which our information-theoretic construction builds.

{\bf Constrained BO.}%
A classical approach to constrained BO is to combine an improvement-based acquisition function with the posterior probability of feasibility, thereby discouraging evaluations that are unlikely to satisfy unknown constraints \citep{schonlau1998global,Gardner2014-Bayesian}. 
More recent information-based approaches directly quantify information about the feasible optimizer or optimum \citep{Hernandez-Lobato2015-Predictive}. 
In particular, the variational information lower bound of \citet{Takeno2022-Sequential} provides a basis for incorporating feasibility into information-theoretic BO and is closely related to the variational construction used in our method.

{\bf Multi-fidelity BO.}
Multi-fidelity BO exploits inexpensive but correlated approximations to reduce the number of costly target-fidelity evaluations. 
Classical approaches include heuristic multi-fidelity extensions of EI such as MFSKO \citep{Huang2005-Sequential}, while subsequent methods developed principled fidelity-selection strategies based on confidence bounds \citep{Kandasamy2017-Multi} or information gain per unit cost, such as multi-fidelity max-value entropy search \citep{Takeno2020-Multifidelity}.
Our method follows the latter information-theoretic perspective, but targets uncertainty about a Pareto frontier rather than a scalar optimum.

{\bf Multi-fidelity multi-objective BO.}
%
MF-OSEMO selects a pair of input and fidelity using output-space entropy reduction about the target (simplified) Pareto frontier \citep{Belakaria2020-multi}, and MF-HVKG uses a one-step hypervolume knowledge-gradient criterion \citep{daulton23-hypervolume}. 
These methods address multiple objectives and fidelities but do not incorporate unknown constraints.

{\bf Constrained multi-fidelity multi-objective BO.} MFMO-SEGO \citep{Charayron2023-towards} addresses constrained multi-fidelity multi-objective optimization by combining a constrained multi-objective acquisition strategy with a separate fidelity-selection criterion. MF-CMOBO \citep{Lin2024-multi} addresses the same setting through a multi-stage procedure combining feasibility search, weighted multi-fidelity hypervolume improvement, and constraint-boundary refinement.
%
%
MF-JESMOC \citep{FernandezSanchez2025-joint} instead provides a unified information-theoretic criterion based on the joint entropy of the feasible Pareto set and frontier.
However, it employs computationally intensive MF-DGPs for every objective and constraint and additional variational inference for Pareto conditioning, whereas our formulation relies on standard GP inference with cross-fidelity covariance to propagate target-fidelity Pareto-consistency information.

\begin{figure*}[t] 

 \centering

 \subfloat[][DTLZ $(L,C,d) = (3,1,7)$]{
 \ig{.3}{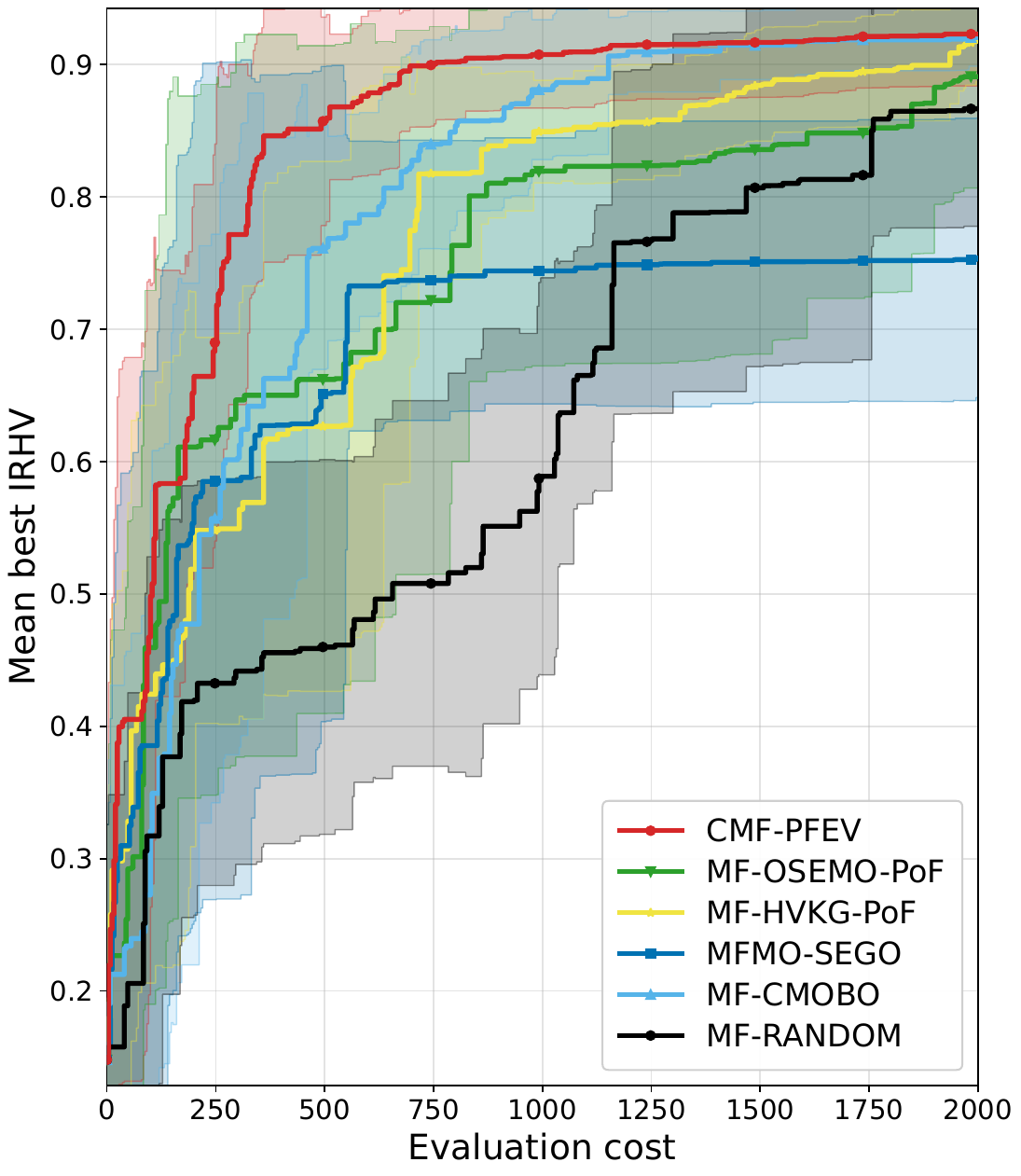}}
 \subfloat[][DTLZ $(L,C,d) = (5,1,9)$]{
 \ig{.3}{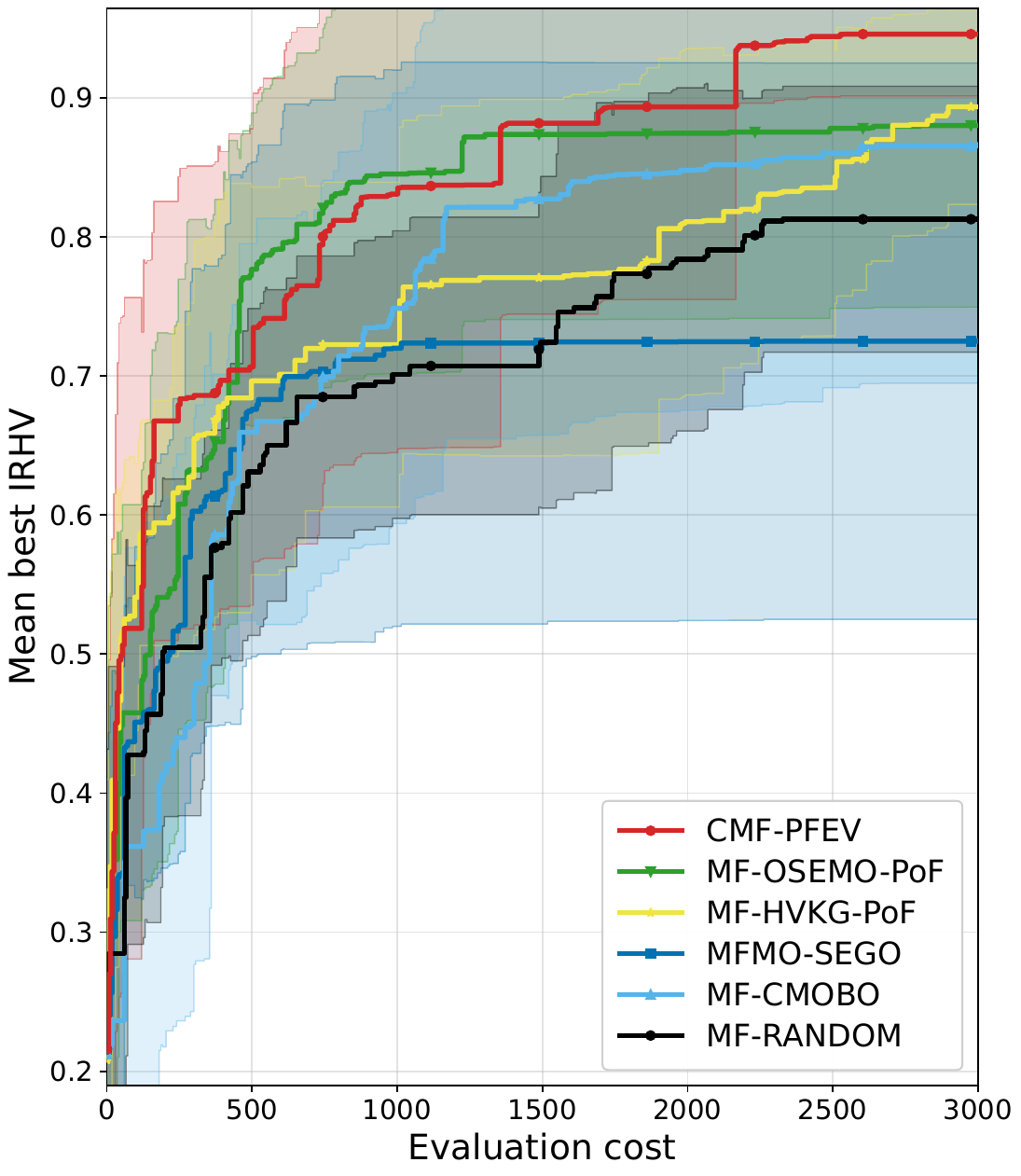}}
 \subfloat[][Tanaka $(L,C,d) = (2,2,2)$]{
 \ig{.3}{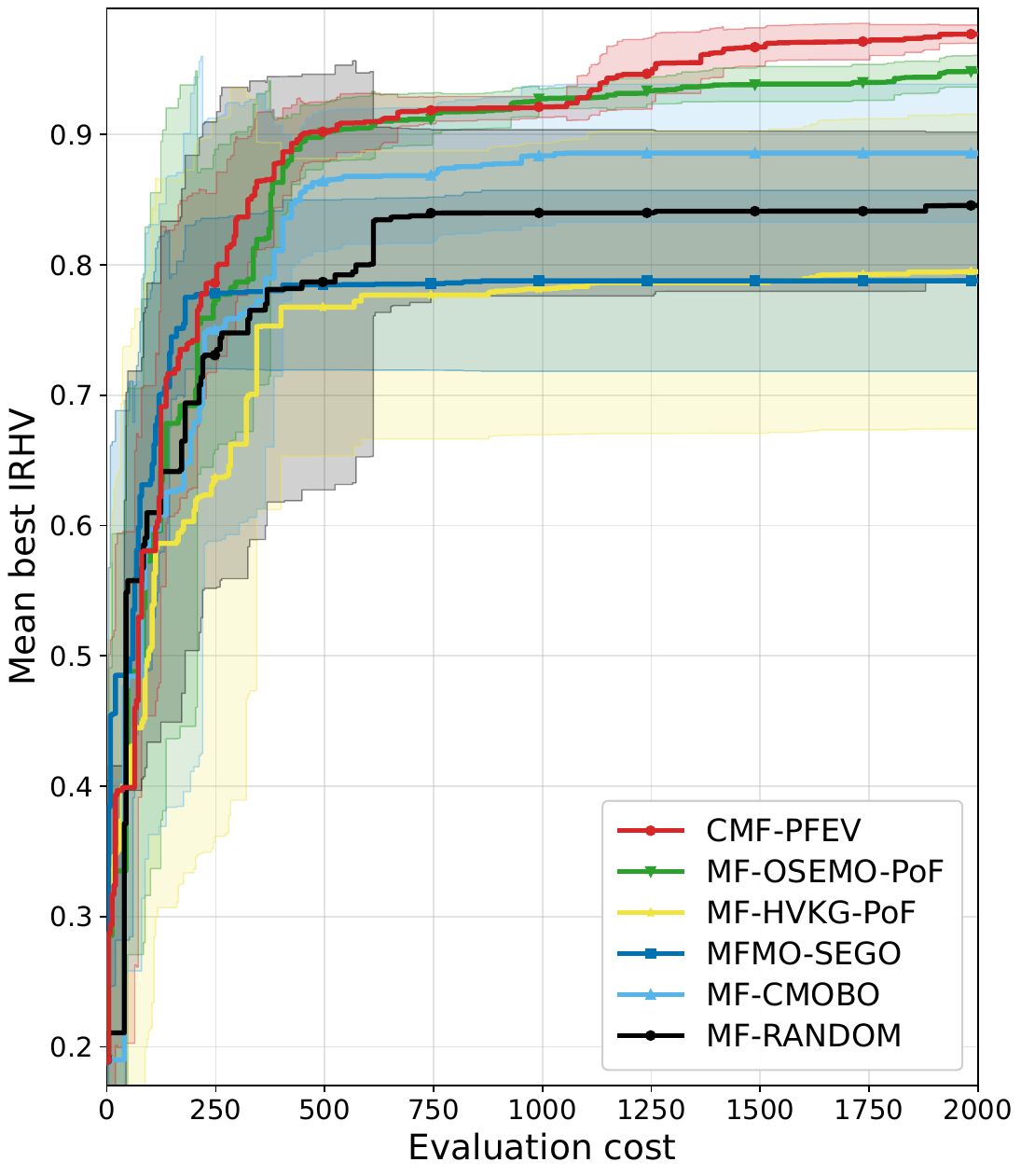}}

 \caption{Results on benchmark functions (10 runs average and standard error).}
 \label{fig:result-bench}

\end{figure*}

\begin{figure*}[t] 

 \centering

 \subfloat[][Waveform $(L,C,d) = (2,1,3)$]{
 \ig{.3}{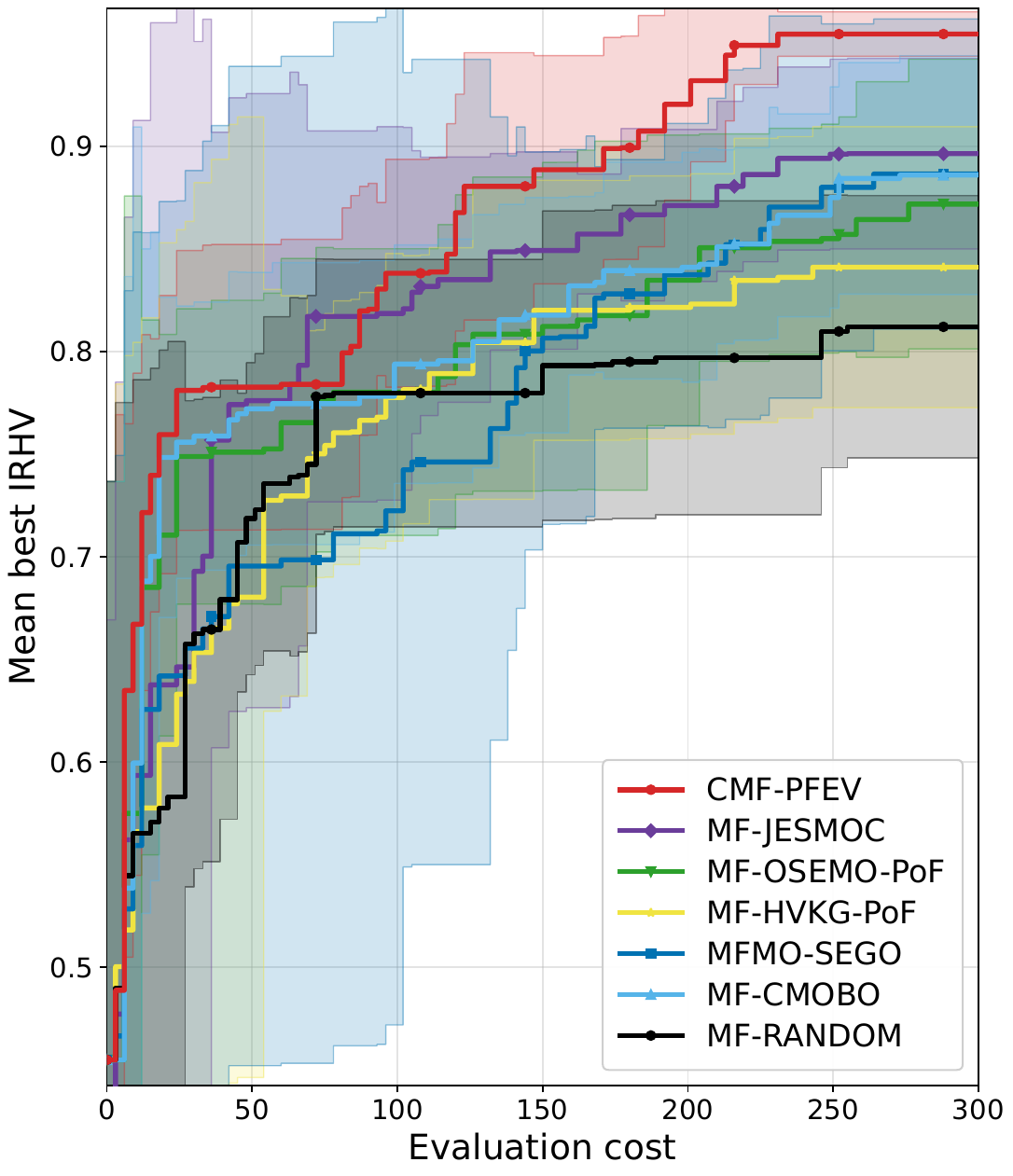}}
 \subfloat[][Bean $(L,C,d) = (2,1,4)$]{
 \ig{.3}{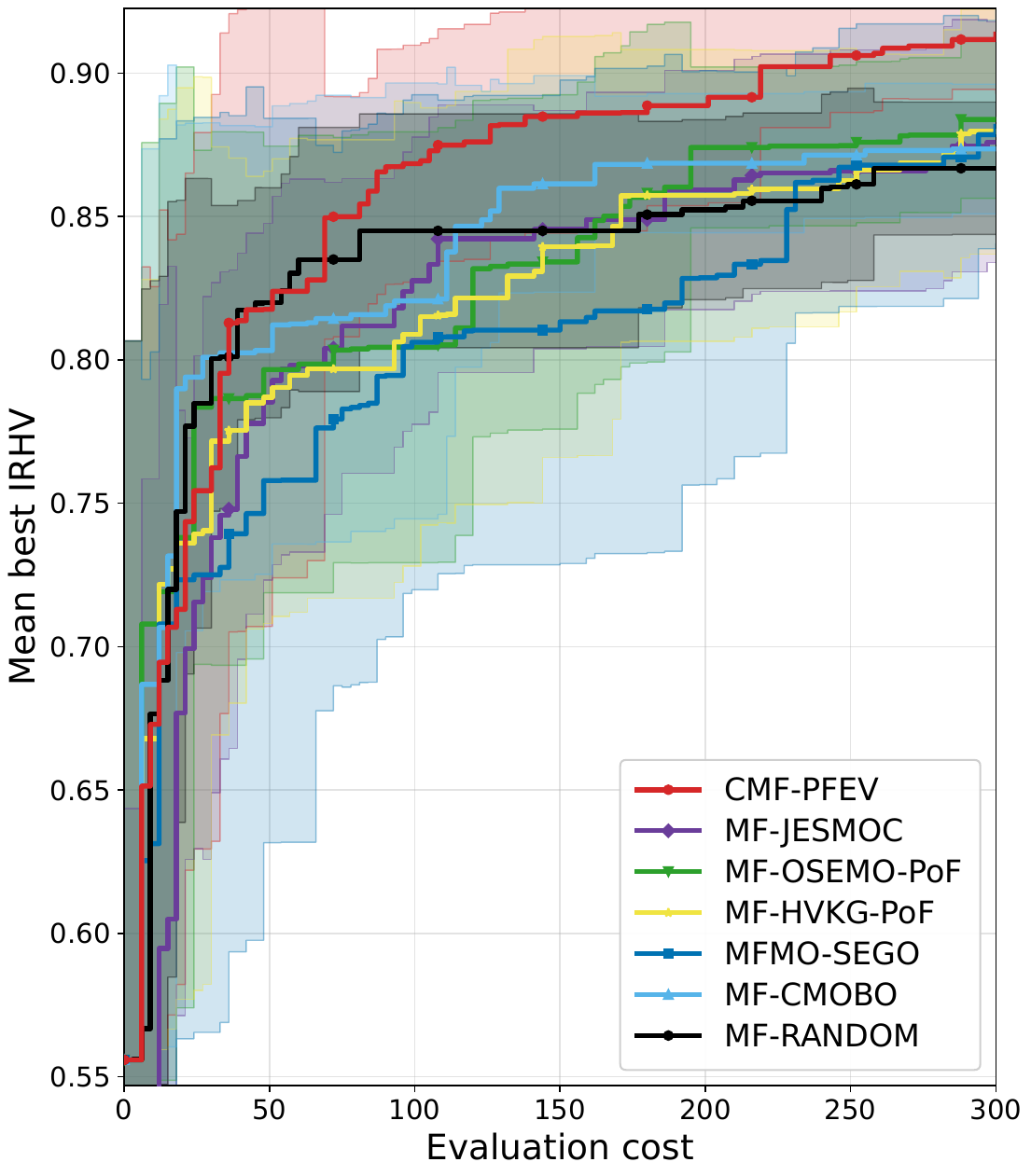}}
 \subfloat[][Covertype $(L,C,d) = (2,1,4)$]{
 \ig{.3}{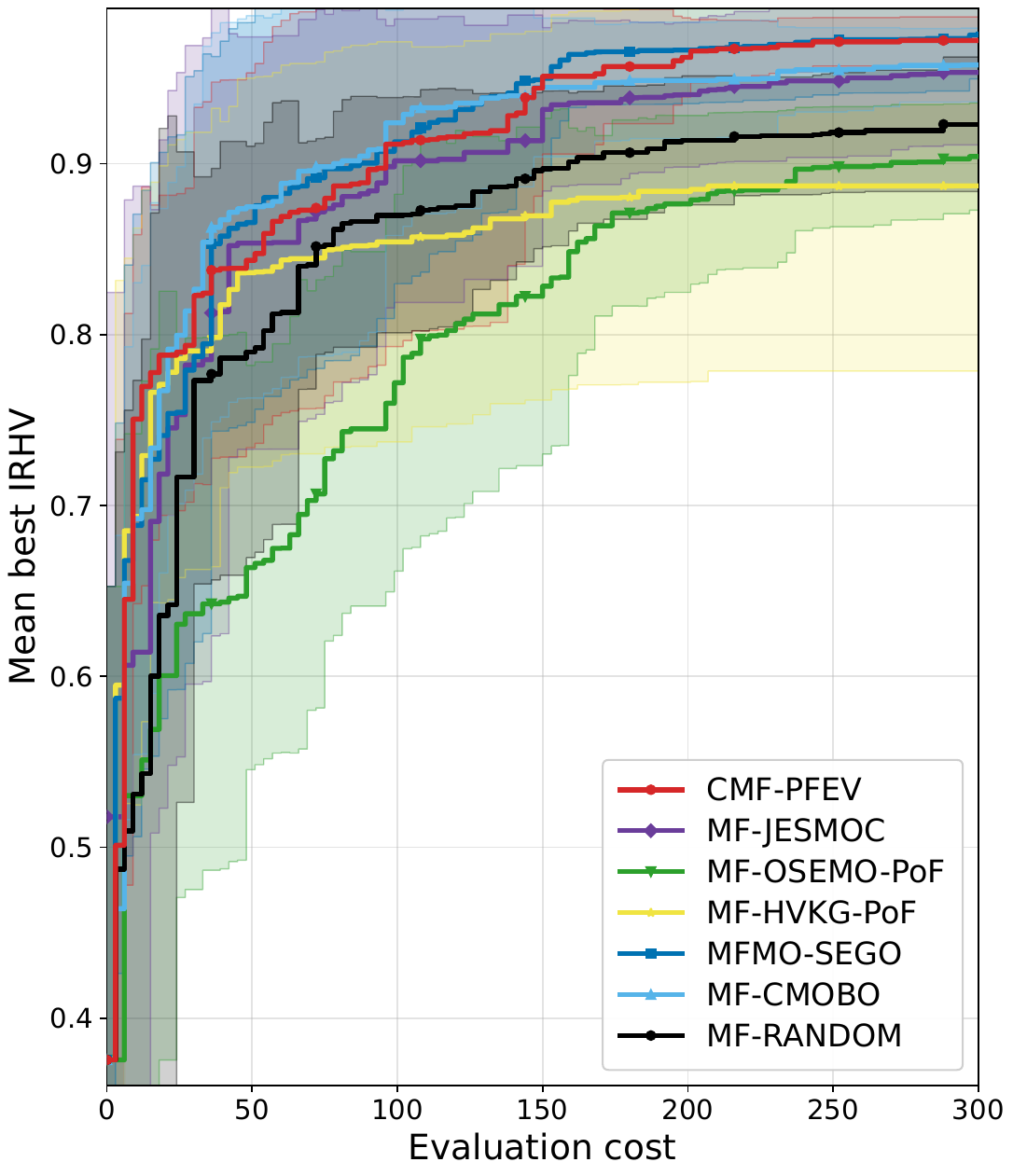}}

 \caption{Results on hyper-parameter selection of LightGBM (10 runs average and standard error).}
 \label{fig:result-automl}

\end{figure*}

\section{Experiments}
\label{sec:experiments}

We evaluate CMF-PFEV on synthetic functions, benchmark functions, and machine-learning hyperparameter optimization problems.
We compare against MF-OSEMO, MF-HVKG, MFMO-SEGO, MF-CMOBO, and MF-JESMOC.
Since MF-OSEMO and MF-HVKG do not handle constraints, we multiply their acquisition functions by the probability of feasibility, following constrained EI
\citep{schonlau1998global,gardner14bayesian};
we denote these variants by MF-OSEMO-PoF and MF-HVKG-PoF.
%
Due to its high computational cost, MF-JESMOC is not shown in some results (see Appendix~\ref{app:computational-time} for time measurements).

We use a product kernel
$k_{\mathrm{Joint}}((\*x,m),(\*x',m')) = k^{(\mathrm{Mat})}(\*x,\*x') k^{(\mathrm{Gau})}(m,m')$,
where 
$k^{(\mathrm{Mat})}$ 
is a Mat\'ern-$5/2$ kernel and
$k^{(\mathrm{Gau})}$ 
is a Gaussian fidelity kernel.
Performance is measured by inference relative hypervolume (IRHV): at each iteration, NSGA-II is applied to the posterior mean to obtain recommended solutions, and their hypervolume is normalized by an approximation of the true Pareto hypervolume.
We report IRHV against cumulative evaluation cost.
All experiments use two fidelity levels ($M=2$), with costs $(L + C)$ and $10 (L + C)$ for synthetic and benchmark problems and $(1,2)$ for hyperparameter optimization. 
Further details are given in Appendix~\ref{app:experimental-detail}.

{\bf GP sample path functions. }%
We generate objective and constraint functions from the joint GP
kernel above, using random Fourier features to obtain continuous
sample paths over $\mathcal{X}$.
We consider several combinations of the numbers of objectives $L$,
constraints $C$, and input dimensions $d$.

{\bf Benchmark functions. }%
We consider the constrained multi-objective benchmarks C2-DTLZ2
\citep{jain2014evolutionary} and Tanaka
\citep{tanaka1995ga}.
Since these are originally single-fidelity problems, we construct
low-fidelity functions by adding GP sample paths to the original
functions. 

{\bf ML hyperparameter optimization. }%
We optimize LightGBM on the Waveform, Dry Bean, and Covertype
datasets. Here, we only describe a setting for Covertype.
The two objectives are macro-F1 and
$-\log_{10}(\text{model size})$, with a constraint on the worst
class-wise recall. The optimized hyperparameters include learning
rate, number of leaves, minimum child samples, and the weight balance
between minority and other classes.
The number of trees is $100$ and $200$ for low and high fidelities, respectively.
Details are in 
Appendix~C.3. 

Fig.~\ref{fig:result-gp-data}, \ref{fig:result-bench}, and \ref{fig:result-automl} show the results, respectively.
Overall, CMF-PFEV performs favorably across all three settings.
The advantage is particularly pronounced on the GP sample path
problems, where the data-generating process matches the GP modeling
assumptions, while consistent improvements are also observed on the
benchmark and hyperparameter optimization problems.

\section{Conclusion}

We proposed CMF-PFEV, an information-theoretic approach to constrained multi-fidelity multi-objective Bayesian optimization.
By expressing information about the target-fidelity feasible Pareto frontier through Pareto-consistency events, we derived a tractable variational acquisition function as a unified criterion. 
The resulting computation relies on Pareto-region probabilities and standard cross-fidelity GP conditioning.
Experiments on synthetic, benchmark, and hyperparameter optimization problems demonstrated the effectiveness of the proposed approach across diverse problem settings.
A limitation is unknown approximation error on the information gain.
Although MI is an essential criterion, its estimation accuracy is still required to be investigated.

\clearpage

\ifdefined\ackn
\section*{Acknowledgments}
This work was partially supported by MEXT KAKENHI (25K03182), MEXT Supporting Pioneering Re-search through AI for 1,000 Discovery challenges Program (SPReAD) Japan Grant Number JPMXP1726275243, and Data Creation and Utilization Type Material Research and Development Project (Grant No. JP-MXP1122712807) of MEXT.
\fi
\bibliography{ref}
\biblstyle

\clearpage
\appendix

\ifdefined\arxiv
\begin{center}
 {\Large{\bf 
 Supplementary Appendix of \\
A Unified Information-Theoretic Approach to Constrained Multi-Fidelity Multi-Objective Bayesian Optimization

 \par}}
\end{center}
\fi

\ifdefined\aistats
\thispagestyle{empty}

\onecolumn
\aistatstitle{
 Supplementary Appendix of \\
A Unified Information-Theoretic Approach to Constrained Multi-Fidelity Multi-Objective Bayesian Optimization

}
\fi


\section{Variational Lower Bound}
\label{app:variational-lower-bound}

The following derivation of the lower bound is a standard approach that exploits nonengativity of Kullback-Leibler (KL) divergence:
\begin{align*}
 &\mr{MI}\rbr{\*r^{(m)}_{\*x};\cF^\ast}
 \\
 & =
 \int p(\cF^*) \int p(\*r^{(m)}_{\*x} \mid \cF^*) 
 \log \frac{ p({\*r^{(m)}_{\*x}} \mid \cF^*) }{ p({\*r^{(m)}_{\*x}}) } \mr{d} \*r^{(m)}_{\*x} \mr{d} \cF^* 
 \notag \\ 
 & = \int p(\cF^*) \sbr{ 
 \int p(\*r^{(m)}_{\*x} \mid \cF^*) \log \frac{ q(\*r^{(m)}_{\*x} \mid \cF^*) }{ p(\*r^{(m)}_{\*x}) } \mr{d} \*r^{(m)}_{\*x} \right.
 \notag \\
 & \qquad \left. + D_{\mr{KL}}\rbr{ 
 p(\*r^{(m)}_{\*x} \mid \cF^*) \parallel q(\*r^{(m)}_{\*x} \mid \cF^*) 
 } 
 } \mr{d} \cF^* 
 \notag \\ 
 &=\EE_{\cF^\ast,\*r^{(m)}_{\*x}}
 \sbr{
   \log
   \frac{q(\*r^{(m)}_{\*x}\mid\cF^\ast)}
        {p(\*r^{(m)}_{\*x})}
 }
 \notag\\
 &\quad
 +\EE_{\cF^\ast}
 \sbr{
   D_{\mr{KL}}\rbr{
   p(\*r^{(m)}_{\*x}\mid\cF^\ast)
   \parallel
   q(\*r^{(m)}_{\*x}\mid\cF^\ast)}
 }
 \\
 &
 \geq
 \EE_{\cF^\ast,\*r^{(m)}_{\*x}}
 \sbr{
   \log
   \frac{q(\*r^{(m)}_{\*x}\mid\cF^\ast)}
        {p(\*r^{(m)}_{\*x})}
 }.
\end{align*}
This indicates that the maximization of the variational lower bound is equivalent to the minimization of the KL divergence
$\EE_{\cF^\ast}
 \sbr{
   D_{\mr{KL}}\rbr{
   p(\*r^{(m)}_{\*x}\mid\cF^\ast)
   \parallel
   q(\*r^{(m)}_{\*x}\mid\cF^\ast)}
 }$.

\section{Overall Procedure}
\label{app:procedure}

Let $B_t$ denote the remaining evaluation budget after the initial data and the first $t$ queries.
Algorithm~\ref{alg:cmf_pfev} summarizes CMF-PFEV. For each fidelity $m\in\{1,\ldots,M\}$, the candidate input $\*x\in\cX$ and mixture weight $\lambda\in(0,1]$ are optimized jointly; the resulting cost-normalized acquisition values are then compared across fidelities. The sampled function paths and observation-noise base samples are held fixed throughout this joint optimization, providing common random numbers across candidate inputs and fidelities. Experimental query-selection rules and numerical settings are specified in Appendix~\ref{app:experimental-detail}.

\begin{algorithm}[tbp]
  \caption{Pseudocode of CMF-PFEV}
  \label{alg:cmf_pfev}
  \small
  \begin{algorithmic}[1]
  \REQUIRE Initial data $\cD_0$, remaining budget $B_0$, costs $\{c_m\}_{m=1}^{M}$, $N_{\mathrm{path}}$, $N_{\epsilon}$, and nominal frontier size $N_{\mathrm{PF}}$
  \STATE \textbf{Function} CMF-PFEV($\cD_0$):
  \STATE $t\leftarrow0$
  \WHILE{$B_t>0$}
    \STATE Update the multi-fidelity GP posterior using $\cD_t$
    \STATE $\mathcal{S}\leftarrow\emptyset$
    \FOR{$k=1,\ldots,N_{\mathrm{path}}$}
      \STATE Draw a posterior function path $\widetilde{\*h}_k$
      \STATE Obtain a finite target-fidelity frontier approximation $\widetilde{\cF}_k^\ast$ with nominal size $N_{\mathrm{PF}}$
      \STATE Draw independent $\*\xi_{k,j}\sim\cN(\*0,\*I_{L+C})$, $j=1,\ldots,N_{\epsilon}$
      \STATE $\mathcal{S}\leftarrow\mathcal{S}\cup\{(\widetilde{\*h}_k,\widetilde{\cF}_k^\ast,\{\*\xi_{k,j}\}_{j=1}^{N_{\epsilon}})\}$
    \ENDFOR
    \STATE Prepare the objective-region cell decompositions
    \FOR{$m=1,\ldots,M$}
      \STATE Maximize over input and mixture weight:
      \STATE $\begin{aligned}
             &(\*x_m,\lambda_m)\leftarrow\\
             &\quad\argmax_{\substack{\*x\in\cX\\\lambda\in(0,1]}}
             \textsc{CalcAcq}(\*x,m,\lambda;\mathcal{S})
             \end{aligned}$
      \STATE $a_m\leftarrow\textsc{CalcAcq}(\*x_m,m,\lambda_m;\mathcal{S})$
    \ENDFOR
    \STATE $m_{t+1}\leftarrow\argmax_{m\in\{1,\ldots,M\}}a_m$ and $\*x_{t+1}\leftarrow\*x_{m_{t+1}}$
    \IF{$c_{m_{t+1}}>B_t$}
      \STATE \textbf{break} \COMMENT{Terminate without observation}
    \ENDIF
    \STATE Observe $\*r_{\*x_{t+1}}^{(m_{t+1})}$ and update $\cD_{t+1}$
    \STATE $B_{t+1}\leftarrow B_t-c_{m_{t+1}}$ and $t\leftarrow t+1$
  \ENDWHILE
  \STATE \textbf{Function} \textsc{CalcAcq}($\*x,m,\lambda;\mathcal{S}$):
  \FOR{$k=1,\ldots,N_{\mathrm{path}}$}
    \STATE Evaluate $Z_{T,k}$ for $T\in\{O,U\}$ using Gaussian CDFs
    \FOR{$j=1,\ldots,N_{\epsilon}$}
      \STATE $\widetilde{\*\epsilon}_{k,j}^{(m)}\leftarrow(\*\Sigma_{\epsilon}^{(m)})^{1/2}\*\xi_{k,j}$
      \STATE $\widetilde{\*r}_{\*x,k,j}^{(m)}\leftarrow\widetilde{\*h}_{\*x,k}^{(m)}+\widetilde{\*\epsilon}_{k,j}^{(m)}$
      \STATE Evaluate $Z_{T,k}(\cdot\mid\widetilde{\*r}_{\*x,k,j}^{(m)})$ for $T\in\{O,U\}$ using Gaussian CDFs
    \ENDFOR
  \ENDFOR
  \STATE Form $\widehat{\cL}_{\mathrm{CMF}}(\*x,m,\lambda)$ using Eq.~\eqref{eq:outer_mc_lb}
  \RETURN $c_m^{-1}\widehat{\cL}_{\mathrm{CMF}}(\*x,m,\lambda)$
  \end{algorithmic}
\end{algorithm}

\section{Experimental Detail}
\label{app:experimental-detail}













We describe the settings used in Section~\ref{sec:experiments}. Continuous inputs are normalized to $[0,1]^d$, whereas machine-learning problems use fixed candidate pools.

\paragraph{Computing infrastructure.}
All experiments were run on an internal compute cluster. The next-query selection times in Table~\ref{tab:runtime_l2_n100_revised} were measured using one CPU thread on an Intel Xeon Gold 5317 processor.

\paragraph{Software and dataset attribution.}
The augmented-input GP models and the MF-HVKG implementation use BoTorch \citep{balandat2020botorch} and GPyTorch \citep{gardner2018gpytorch}. The classifier experiments use LightGBM \citep{ke2017lightgbm}. Original comparison methods and benchmark functions are cited in their respective descriptions below. Waveform-5000 is derived from the \url{https://doi.org/10.24432/C56014}{Waveform Database Generator (Version 2)} dataset of Breiman and Stone (1984); \url{https://doi.org/10.24432/C50S4B}{Dry Bean} is due to K\"okl\"u and \"Ozkan (2020); and \url{https://doi.org/10.24432/C50K5N}{Covertype} is due to Blackard (1998). The UCI Machine Learning Repository lists these three source datasets under the Creative Commons Attribution 4.0 International (CC BY 4.0) license. They contain synthetic waveforms, bean measurements, and forest-cover attributes, respectively; we do not collect participant data.


\paragraph{Initial data and budget.}
Each run starts from five distinct inputs evaluated at both fidelities for every objective and constraint. Thus, each objective or constraint GP initially has ten observations. Inputs use scrambled Sobol sampling for continuous problems and uniform sampling without replacement for pools. Methods share initial data for each of ten BO seeds, $0,\ldots,9$. The test functions, data splits, and classifier seeds remain fixed across these runs.

A coupled query costs $c_m=(L+C)\kappa_m$, with per-output costs $(\kappa_1,\kappa_2)=(1,10)$ for synthetic and benchmark problems and $(1,2)$ for machine-learning problems. The additional budget after initialization is $50c_2$. Figures start at zero by subtracting the initial cost $5(c_1+c_2)$; the initial observations remain available to all methods.

\paragraph{Experimental query-selection rules.}
In all experiments, after ten consecutive low-fidelity queries, the next query is restricted to the highest fidelity. Its input is selected by the method's own rule at that fidelity, and a highest-fidelity query resets the counter. For pool problems, previously observed input--fidelity pairs are excluded from the candidate set; an input observed at one fidelity remains available at another fidelity if that pair is unobserved. These rules are applied to all compared methods.

\paragraph{GP model.}
Except for MF-JESMOC's deep GP, methods use independent augmented-input GPs with the product kernel in Section~\ref{sec:experiments}. Its factors are
\begin{align*}
 k^{(\mathrm{Mat})}(\boldsymbol{x},\boldsymbol{x}')
 &=\sigma_f^2
 \left(1+\sqrt{5}\,r+\frac{5r^2}{3}\right)e^{-\sqrt{5}\,r},
 & r^2&=\sum_{i=1}^{d}\frac{(x_i-x'_i)^2}{\ell_i^2},\\
 k^{(\mathrm{Gau})}(m,m')
 &=\exp\left[-\frac{(s_m-s_{m'})^2}{2\ell_s^2}\right],
 & s_m&=\frac{m-1}{M-1}.
\end{align*}
The Mat\'ern-$5/2$ kernel uses ARD: a separate input lengthscale $\ell_i$ is learned for each dimension. Each output has its own constant mean, kernel parameters, and noise variance. Outputs are standardized separately. Hyperparameters are refitted after each query by marginal-likelihood maximization, warm-started from the preceding fit, with up to eight attempts. In standardized units, the bounds are $[10^{-6},1]$ for noise variance, $[10^{-3},100]$ for lengthscales, and $[10^{-6},100]$ for output variance.

\paragraph{CMF-PFEV sampling and frontiers.}
We use $N_{\mathrm{path}}=10$ posterior paths and $N_{\epsilon}=4$ observation-noise draws per path in Eq.~\eqref{eq:outer_mc_lb}. Prior paths use random Fourier features drawn from the input and fidelity kernels' spectral distributions. The two feature maps are combined by a tensor product, giving $45\times46=2070$ features from a nominal budget of $2048$. A Matheron-style correction conditions each path on the observations using the fitted kernel covariance and simulated training noise. GP fitting uses the kernel covariance directly.

For each sampled target-fidelity path, constrained NSGA-II uses population $50$ and $1000$ generations, with nominal frontier size $N_{\mathrm{PF}}=50$. For pools, all candidates are evaluated on the sampled path, and the sampled-feasible nondominated set is extracted. Sets larger than $50$ are pruned using the crowding-distance criterion of NSGA-II, retaining objective-wise extremes and favoring sparsely represented regions. If no sampled-feasible point is found, least-violating points provide a numerical fallback.

\paragraph{Acquisition optimization.}
Both objective-region and feasibility probabilities are evaluated using Gaussian CDFs; the objective calculation uses disjoint cell decomposition as in Section~\ref{sec:cmf_exact_prob}. For each fidelity, continuous inputs and $\lambda\in[10^{-4},1]$ are jointly optimized by L-BFGS-B with four restarts, $128$ raw initialization samples, and at most $100$ iterations. For pools, each available input is fixed while $\lambda$ is optimized continuously. Each sampled pool frontier retained without pruning uses $\lambda_k=0$ in its summand of Eq.~\eqref{eq:outer_mc_lb}; if all sampled frontiers are retained in full, $\lambda$ is fixed to zero throughout acquisition optimization. The best cost-normalized pair is selected subject to the experimental query-selection rules above. MF-OSEMO-PoF also uses four restarts, $128$ raw initialization samples, and at most $100$ iterations for continuous acquisition optimization; MFMO-SEGO uses the constrained optimizer described below.

\paragraph{Comparison methods.}
MF-OSEMO-PoF and MF-HVKG-PoF use highest-fidelity feasibility:
\begin{equation*}
 \operatorname{PoF}_M(\boldsymbol{x})
 =\prod_{c=1}^{C}\Phi\left(
 \frac{\mu_{g,c}^{(M)}(\boldsymbol{x})-z_c}
 {\sigma_{g,c}^{(M)}(\boldsymbol{x})}\right),
\end{equation*}
where constraint means and standard deviations are in the original units.

MF-OSEMO-PoF uses the numerical-integration variant of MF-OSEMO \citep{Belakaria2020-multi}, ten frontier samples with the above path sampler, and observation noise. Our implementation uses $64$ Gauss--Hermite nodes for lower-fidelity entropy and an analytic highest-fidelity term.

MF-HVKG-PoF uses BoTorch's one-shot implementation \citep{daulton23-hypervolume} with $32$ fantasy observation scenarios and $10$ target-fidelity recommendation points per scenario. Utility is the hypervolume of the updated posterior means. For each discrete fidelity, the query input and fantasy-specific recommendation points are jointly optimized, initializing the latter from the optimized current recommendation set. Acquisition optimization uses one restart, $128$ raw initialization samples, and at most $200$ iterations; current-value optimization uses one restart, $256$ raw samples, and at most $200$ iterations. For pools, query inputs are enumerated and each fantasy's recommendation set is reoptimized within the pool by hypervolume-increasing single-point exchanges, with at most $10$ sweeps. The current posterior-mean frontier, pruned to at most $10$ points by crowding distance, supplies the initialization.

MFMO-SEGO selects an input by maximizing WB2S-MPI subject to highest-fidelity posterior-mean constraints, $\mu_{g,c}^{(M)}(\boldsymbol{x})\geq z_c$ for all $c$ \citep{Charayron2023-towards}. Its posterior-mean frontier uses NSGA-II with population $50$ and $1000$ generations, or pool enumeration, retaining all resulting frontier points. WB2S-MPI combines the minimum probability of improvement against these points with a smooth minimum of normalized objective means, using temperature $0.05$ and $\beta=100$. The MPI scaling reference is optimized over the same mean-feasible domain; the raw acquisition is used without PoF weighting or nonnegative shifting. Continuous optimization uses SLSQP with four restarts, $128$ Sobol initialization samples supplemented by posterior-mean Pareto inputs, and at most $100$ iterations, with feasibility tolerance $10^{-7}$; pools are enumerated. If no mean-feasible candidate is found, a feasibility-restoration safeguard selects a candidate with minimum total predicted constraint violation. After selecting the input, each objective chooses the fidelity maximizing the Gaussian conditional reduction in target-fidelity variance, including observation noise, divided by squared cost. The highest of these fidelities is used for the coupled query.

MF-CMOBO combines feasibility search, weighted multi-fidelity expected hypervolume improvement, and constraint-boundary refinement \citep{Lin2024-multi}. Improvement and boundary refinement alternate after feasibility search. Entropy weights use $5000d$ Latin-hypercube locations, and cross-fidelity scaling uses $1000d$ test locations, or all pool candidates. Constraint relaxation starts at $-3$ and increases by $1.5$ up to $6$. Continuous optimization uses eight restarts, $128$ raw samples, and at most $100$ iterations, with a smooth constraint gate of temperature $0.05$ followed by a hard feasibility check; pools are enumerated.

MF-JESMOC \citep{FernandezSanchez2025-joint} uses the authors' multi-fidelity deep-GP implementation, restricted to $L\leq2$ because of its computational cost. Its two training phases use full batches, $5000/15000$ iterations, and learning rates $0.003/0.001$. We retain the authors' defaults of $50$ Pareto-set points and $1000d$ optimization-grid locations.

MF-RANDOM samples fidelity uniformly among allowed levels and uses scrambled Sobol inputs or uniform sampling among available pool candidates.

\paragraph{Evaluation.}
IRHV uses posterior-mean recommendations, obtained by constrained NSGA-II with population $300$ and $1000$ generations or pool enumeration. Recommendations are scored on noise-free highest-fidelity functions, discarding truly infeasible points; these evaluations do not enter BO data or consume its budget. Continuous reference frontiers use population $300$ and $10000$ generations, whereas pool references are exact. Each problem has a fixed hypervolume reference vector and denominator shared by all methods and seeds. Figures average the running-best IRHV of the ten seeds, with bands of $\pm$ one standard error. Scores use the latest completed query at each cost and are carried forward to the budget limit.

\subsection{GP sample path functions}
\label{app:gp-sample-functions}

Each objective and constraint is an independent fixed prior draw from the product kernel above, with mean zero, variance one, isotropic input lengthscale $\ell_x$, and fidelity lengthscale $\ell_s=5$. Table~\ref{tab:app-gp-settings} lists the six settings. Generation uses $4096$ random Fourier features, each with a joint input--fidelity frequency drawn from the Mat\'ern-$5/2$ and RBF spectral distributions. Independent Gaussian observation noise has standard deviation $0.01$ at both fidelities, and constraint thresholds are zero.

\begin{table}[htbp]
\centering
\caption{GP-generated test-function settings. Each problem is fixed across BO seeds.}
\label{tab:app-gp-settings}
\begin{tabular}{rrrr}
\hline
$L$ & $C$ & $d$ & $\ell_x$ \\
\hline
2 & 1 & 2 & 0.20 \\
2 & 1 & 5 & 0.25 \\
2 & 2 & 2 & 0.20 \\
2 & 2 & 5 & 0.25 \\
3 & 1 & 5 & 0.20 \\
3 & 2 & 5 & 0.20 \\
\hline
\end{tabular}
\end{table}

\subsection{Benchmark functions}
\label{app:benchmark-functions}

The highest-fidelity functions are Tanaka \citep{tanaka1995ga} with $(L,C,d)=(2,2,2)$ and C2-DTLZ2 \citep{jain2014evolutionary} with $(3,1,7)$ and $(5,1,9)$. Objectives are negated for maximization, and constraints are expressed as nonnegative feasible margins. Tanaka maps normalized inputs to $\boldsymbol{u}=\pi\boldsymbol{x}$ and uses objectives $(-u_1,-u_2)$. C2-DTLZ2 uses constraint radius $0.4$ for $L=3$ and $0.5$ for $L=5$.

For each output $h_j$, an independent RBF-GP discrepancy $b_j$ is generated with lengthscale $0.30$ and $1024$ Fourier features. Let $\overline b_j$, $s_{b,j}$, and $s_{H,j}$ be its mean, its standard deviation, and the highest-fidelity output standard deviation, calculated at $4096$ fixed uniform calibration inputs. Low-fidelity outputs are
\begin{equation*}
 h_j^{(m)}(\boldsymbol{x})
 =h_j^{(M)}(\boldsymbol{x})
 +0.25(1-s_m)s_{H,j}
 \frac{b_j(\boldsymbol{x})-\overline b_j}{s_{b,j}}.
\end{equation*}
All constraint thresholds are zero. Independent Gaussian observation noise has standard deviation $0.01$ at both fidelities.

\subsection{LightGBM hyperparameter selection}
\label{app:light-gbm}

\paragraph{Datasets.}
Waveform-5000 uses OpenML ID $60$, Dry Bean uses UCI ID $602$, and Covertype uses scikit-learn's dataset loader. Fixed stratified training/validation splits are $50{:}50$ for Waveform and $70{:}30$ for Dry Bean. Covertype selects $60000$ training and $40000$ validation examples from $581012$ examples (Table~\ref{tab:app-ml-data}). All input columns are numerical; missing values are imputed using training-set medians.

\begin{table}[htbp]
\centering
\caption{LightGBM dataset settings. All use $512$ candidates, $L=2$, and $C=1$.}
\label{tab:app-ml-data}
\begin{tabular}{lrrrrr}
\hline
Dataset & Classes & Features & Train & Validation & $d$ \\
\hline
Waveform-5000 & 3 & 40 & 2500 & 2500 & 3 \\
Dry Bean & 7 & 16 & 9527 & 4084 & 4 \\
Covertype & 7 & 54 & 60000 & 40000 & 4 \\
\hline
\end{tabular}
\end{table}

\paragraph{Hyperparameters and fidelities.}
Each pool contains $512$ fixed scrambled Sobol inputs. Coordinates map logarithmically to learning rate $[0.01,0.20]$, number of leaves $[7,127]$, and minimum samples per leaf $[2,50]$ for Waveform or $[2,100]$ for Dry Bean and Covertype. A coordinate $x_i\in[0,1]$ maps to $a(b/a)^{x_i}$ for range $[a,b]$; integer parameters are rounded. Dry Bean and Covertype add class-weight ratio $\rho\in[0.1,10]$. The rare group consists of the least frequent $\lfloor K_{\mathrm{class}}/2\rfloor$ training classes (encoded labels break ties). Class weights are $\rho$ for rare classes and one otherwise, normalized by their mean across classes.

Low and high fidelities use $100$ and $200$ boosting rounds. LightGBM uses deterministic GBDT training with a fixed classifier seed, no row/feature subsampling, and no $L_1/L_2$ regularization. Candidate outputs are precomputed and returned from the saved tables during BO, without added observation noise.

\paragraph{Objectives and constraints.}
Waveform and Covertype maximize validation macro-F1 and the negative base-10 logarithm of model size in bytes, measured from the booster's UTF-8 serialization. The constraint metric is minimum class-wise recall. Dry Bean maximizes macro-average recall in its rare and common groups (three and four classes), with validation accuracy as the constraint metric.

For metric $a^{(m)}$, BO observes $g^{(m)}=a^{(m)}-\theta$ with threshold zero. The fixed $\theta$ is the $0.70$ quantile of full-pool highest-fidelity values for Waveform/Covertype and the $0.50$ quantile for Dry Bean; it applies at both fidelities (Table~\ref{tab:app-ml-thresholds}). The optimizer receives only queried outputs and this threshold. For IRHV evaluation, posterior-mean pool recommendations are pruned by crowding distance to the exact feasible frontier sizes in the table. This evaluation rule applies to every method; these sizes are not used by acquisition optimization.

\begin{table}[htbp]
\centering
\caption{Constraint thresholds and feasible highest-fidelity pool frontier sizes. Stored full-precision thresholds are used in calculations.}
\label{tab:app-ml-thresholds}
\begin{tabular}{llrrr}
\hline
Dataset & Constraint metric & Quantile & $\theta$ & Frontier size \\
\hline
Waveform-5000 & Worst-class recall & 0.70 & 0.7892434988 & 6 \\
Dry Bean & Accuracy & 0.50 & 0.9269098923 & 10 \\
Covertype & Worst-class recall & 0.70 & 0.5836391437 & 13 \\
\hline
\end{tabular}
\end{table}

\section{Computational Time}
\label{app:computational-time}

Table~\ref{tab:runtime_l2_n100_revised} shows computational time comparison on the GP data.
%
MFMO-SEGO and MF-CMOBO, which are based on separated heuristic approach, are relatively faster. 
%
On the other hand, information-based and knowledge gradient based approaches are slow compared with them.
%
MF-JESMOC obviously took longer times. 
This is due to the estimation of complicated deep GP and variational approximations.

\begin{table}[t]
\centering
\caption{
Next-query selection time on a GP-sampled synthetic problem with $L=2$ objectives, $C=2$ constraints, $d=5$, $M=2$ fidelity levels, and $N=100$ total observations (50 at each fidelity). Values are the median and quartiles over five seeds, measured with one CPU thread (Xeon Gold 5317).}
\label{tab:runtime_l2_n100_revised}
\begingroup
\small
\setlength{\tabcolsep}{3pt}
\begin{tabular}{lrr}
\toprule
Method & Median (s) & $[Q_1,Q_3]$ (s) \\
\midrule
CMF-PFEV & 103.1 & $[102.5,\,121.6]$ \\
MF-OSEMO-PoF & 50.2 & $[50.0,\,50.2]$ \\
MF-HVKG-PoF & 26.6 & $[13.7,\,30.9]$ \\
MFMO-SEGO & 13.2 & $[13.0,\,13.5]$ \\
MF-CMOBO & 12.7 & $[11.5,\,12.8]$ \\
MF-JESMOC & 3354.1 & $[3339.6,\,3364.0]$ \\
\bottomrule
\end{tabular}
\endgroup
\end{table}
\end{document}